\documentclass[conference]{IEEEtran}
\IEEEoverridecommandlockouts
\usepackage[utf8]{inputenc}
\usepackage[T1]{fontenc}
\usepackage[hyphens]{url}
\usepackage[breaklinks, colorlinks, linkcolor=black, anchorcolor=black, citecolor=black, urlcolor=black]{hyperref}
\usepackage{amsmath,graphicx,makecell}
\usepackage{enumitem} 
\graphicspath{{./fig/}}
\usepackage{threeparttable}
\usepackage{booktabs}
\usepackage{algorithmic}
\usepackage[linesnumbered, ruled, vlined]{algorithm2e}
\usepackage{adjustbox}
\usepackage{booktabs}
\usepackage{amssymb}
\usepackage{wrapfig}
\usepackage{multicol}         
\usepackage{multirow}         
\usepackage{url}
\usepackage{tabularx}
\usepackage{xurl}
\usepackage{cite}
\usepackage{amsmath,amssymb,amsfonts,amsthm}
\usepackage{xcolor}
\usepackage{soul} 
\usepackage{algorithmic}
\usepackage{graphicx}
\usepackage{textcomp}
\usepackage{url}
\usepackage{tabularx}
\usepackage{xurl}
\usepackage{threeparttable}
\usepackage{xcolor}
\usepackage{caption}
\usepackage{subcaption}
\usepackage{amsmath,amssymb}

\usepackage{color}

\definecolor{NoteColor}{RGB}{0,104,149}

\usepackage{pifont}
\newcommand{\cmark}{\ding{51}}%
\newcommand{\xmark}{\ding{55}}%

\definecolor{midnightblue}{rgb}{0.1, 0.1, 0.44}

\def\BibTeX{{\rm B\kern-.05em{\sc i\kern-.025em b}\kern-.08em
    T\kern-.1667em\lower.7ex\hbox{E}\kern-.125emX}}
\begin{document}

\title{PowerMamba: A Deep State Space Model and Comprehensive Benchmark for Time Series Prediction in Electric Power Systems\vspace{-2mm}}

\author{
    Ali Menati$^\ast$, Fatemeh Doudi$^\ast$, Dileep Kalathil$^\dagger$, and Le~Xie$^{\ddagger,~\S}$ \\
    \{$^\ast$\IEEEmembership{Student Member}, $^\dagger$\IEEEmembership{Senior Member}, $^\ddagger$\IEEEmembership{Fellow}\},~IEEE

    \thanks{Ali Menati, Fatemeh Doudi, Dileep Kalathil are with the Department of Electrical and Computer Engineering, Texas A\&M University, College Station, TX, USA. Le Xie is with the School of Engineering and Applied Sciences, Harvard University, Boston MA, USA.}
    \thanks{$\S$ Corresponding Author: Le Xie, email: xie@seas.harvard.edu}
}

\IEEEoverridecommandlockouts

\makeatletter
\def\@maketitle{%
  \newpage
  \null
  \vspace{-10mm} 
  \begin{center}%
    {\fontsize{24pt}{28pt}\selectfont 
     \@title\par}%
    \vskip 1.3em 
    {\normalsize \lineskip .5em%
      \begin{tabular}[t]{c}\@author
      \end{tabular}\par}%
    \vskip 1.1em 
  \end{center}%
  \par
  \vspace{-5mm} 
}
\makeatother

\maketitle

\begin{abstract}
The electric power sector is undergoing substantial transformations due to the rising electrification of demand, enhanced integration of renewable energy resources, and the emergence of new technologies. These changes are rendering the electric grid more volatile and unpredictable, making it more challenging to maintain reliable operations. In order to address these issues, advanced time series prediction models are needed for closing the gap between the forecasted and actual grid outcomes. In this paper, we introduce a multivariate time series prediction model that combines traditional state space models with deep learning methods to simultaneously capture and predict the underlying dynamics of multiple time series. Additionally, we design a time series processing module that incorporates high-resolution external forecasts into sequence-to-sequence prediction models, achieving this with negligible increases in size and no loss of accuracy. We also release an extended dataset spanning five years of load, electricity price, ancillary service price, and renewable generation. To complement this dataset, we provide an open-access toolbox that includes our proposed model, the dataset itself, and several state-of-the-art prediction models, thereby creating a unified framework for benchmarking advanced machine learning approaches.  Our findings indicate that the proposed model outperforms existing models across various prediction tasks, improving state-of-the-art prediction error by an average of 7\% and decreasing model parameters by 43\%. The code is
available at {\href{https://github.com/alimenati/PowerMamba}{\color{blue}https://github.com/alimenati/PowerMamba}}

\end{abstract}


\section{Introduction} 
The growing integration of renewable energy resources into the electric grid has introduced substantial uncertainty in generation. Simultaneously, demand electrification, electric vehicles, the widespread deployment of energy storage systems, and the rise of flexible loads, such as computing data centers, are transforming load profiles and making them increasingly challenging to predict \cite{zhang2025unlocking, hussain2023enhancing, hussain2022optimization}. Beyond uncertainties in load and generation, their interactions through electricity market price signals introduce further complexity to grid dynamics \cite{shi2024demand}. Price-responsive resources actively shape market prices while adjusting their operations in response \cite{menati2023high}. These interdependencies necessitate forecasting models that incorporate market behavior and price dynamics to accurately capture the relationships between load, generation, and prices. Accurate forecasting is vital for all grid participants because it lowers system costs, enhances reliability, and enables greater renewable energy integration, advancing the goal of achieving a net-zero electric grid.

While many time series prediction models focus on forecasting load, renewable generation, or market prices separately, there is currently no existing framework for predicting these variables together
while capturing their correlations and interactions \cite{9732470}. The use of
distinct models for each variable reduces accuracy and raises computational costs, since each model demands its own dataset, training process, and inference pipeline. Furthermore, benchmarking these models and adapting them to broader use cases is very challenging \cite{10012043}, highlighting the need for a unified, open-access framework that is capable of supporting multivariate time series prediction, integrating existing models, and keeping up with the rapid advancements in machine learning research and methodology.

Building accurate multivariate prediction models requires
datasets that capture the diverse statistical patterns and complicated dynamics of electric grids. Existing datasets are often aggregated over large areas and lack the spatial granularity that is needed to capture zonal or nodal variation \cite{STLF}. Additionally, they usually have a short duration, which limits their applicability for training sophisticated models such as transformers that perform well with long-term, diverse, high-resolution data \cite{yang2017electricity}. Developing state-of-the-art prediction models requires the creation of large 
datasets with higher spatial granularity, longer temporal coverage, and precise representation of the interactions among grid variables.

Most time series prediction models operate as sequence-to-sequence frameworks that depend only on a window of past observations to predict a corresponding future window \cite{10111057}. However, external factors like weather have a significant impact on power systems, and they cannot be captured by historical records. Hence, publicly available forecasts for certain time series, like load and renewable generation, that are based on climate models and weather data, can offer significant potential to enhance accuracy. These external predictions are frequently overlooked, and the existing models that try to incorporate such predictions either decrease the temporal resolution of future inputs, which compromises accuracy, or increase the number of input features, which inflates the model’s size \cite{fu2023masked}. Developing models that use high-resolution external forecasts effectively without adding complexity is a crucial step in enhancing scalable prediction tools.

\begin{figure}[hbt]
    \centering
    \vspace{-2mm}
    \includegraphics[width=.75\linewidth]{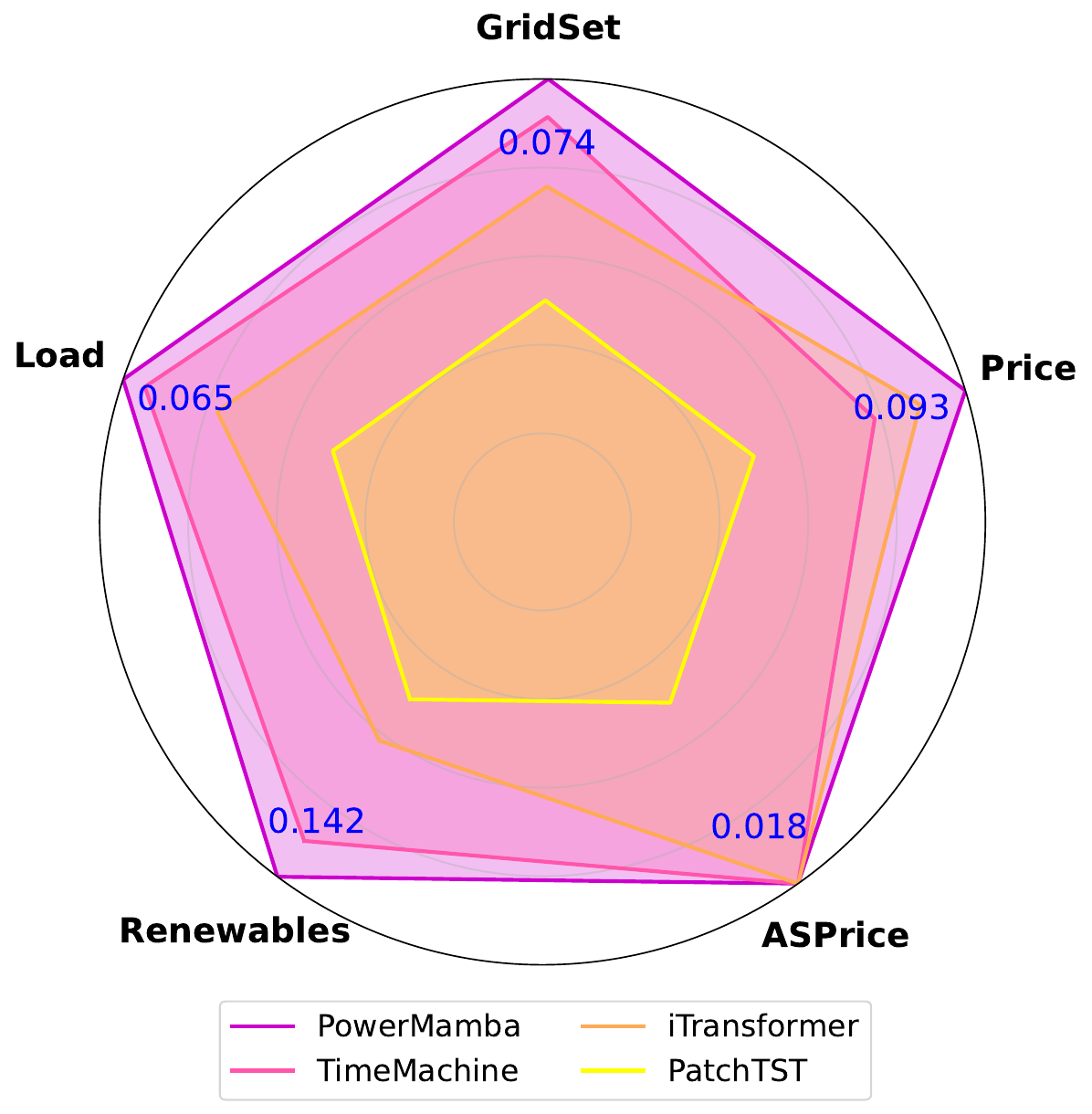}
    \caption{Average Mean Squared Error (MSE) comparison between PowerMamba and state-of-the-art baselines with a context length of 240 hours and a 24-hour prediction window. The circle center represents the maximum possible error, and closer to the boundary indicates better performance.}
    \label{fig:circle}
    \vspace{-5mm}
\end{figure}

Recent advancements in artificial intelligence, especially Transformer-based models, have substantially enhanced accuracy in time series prediction tasks. Nevertheless, transformers are computationally expensive, with complexity scaling quadratically with context length, rendering them inefficient for applications that require long-range dependency modeling \cite{tay2020long}. Deep State Space Models (SSMs) offer a promising alternative by capturing long-term dependencies with linear computational complexity \cite{gu2021efficiently}. In particular, Mamba, a selective SSM architecture, has been able to surpass transformer models in both accuracy and efficiency for a variety of complex tasks \cite{gu2023mamba}. In this paper, we build on recent advancements in selective SSMs to design a multivariate prediction model and release an open-access toolbox that addresses the key challenges outlined in this introduction by providing a comprehensive solution for time series prediction in power systems. Our main contributions are summarized as follows:

\begin{enumerate}

\item We propose PowerMamba, a multivariate forecasting model that combines the efficiency of selective state-space models with architectural innovations tailored to power system time series. PowerMamba uses Mamba blocks to capture the latent dynamics of grid variables such as load, price, and renewable generation, which often follow differential equations. It uses a time series decomposition module that isolates trend and seasonal components and improves performance on highly variable time series like wind and solar. We adopt a dual-path architecture with standard and inverse Mamba blocks, enabling the model to learn both temporal patterns within each time series and cross-channel dependencies among variables. As shown in Figure~\ref{fig:circle}, PowerMamba outperforms current benchmarks in all prediction tasks while reducing parameter count by 43\% compared to the best Mamba-based model and 78\% compared to the leading Transformer-based model.

\item To capture the influence of exogenous factors shaping power system behavior, we introduce a time series processing block that integrates high-resolution external forecasts, such as ERCOT’s day-ahead load and renewable generation predictions. These forecasts, often highly dependent on weather conditions, provide valuable forward-looking signals that complement historical observations. The module substantially improves forecasting accuracy, even for time series without direct external forecast inputs, while maintaining model size and input resolution.

\item We provide an open-access toolbox that includes PowerMamba and other state-of-the-art machine learning models, enabling benchmarking across a wide range of short- and long-term forecasting tasks. To support this, we release a comprehensive  dataset based on ERCOT with five years of hourly data on zonal loads, electricity prices, ancillary service prices, and renewable generation. It includes 22 core time series and an extended version with 262 channels featuring external forecasts, allowing model evaluation on the full dataset or selected subsets.

\end{enumerate} 

 Section~\ref{Literature} presents relevant literature and its relation to our work. Section~\ref{Mamba} introduces deep state space models and Mamba. Our time series forecasting model, PowerMamba, is presented in Section~\ref{PowerMamba}, and our comprehensive dataset is outlined in Section~\ref{dataset}. Section~\ref{results} compares different aspects of our model with selected benchmark models. Finally, Section~\ref{conclusion} concludes the paper and indicates future directions.


\section{Literature Review}
\label{Literature}

This paper introduces a foundational model-based approach and a data benchmark for multivariate time series prediction in electric power systems. Time series forecasting is a core function in many aspects of power system operation and planning \cite{hippert2001neural, menati2022competitive}. Methods of time series prediction in power
systems have evolved from statistical methods to advanced machine learning-based techniques. Early methods such as ARIMA \cite{contreras2003arima} were effective for univariate stationary time
series, with extensions like SARIMA \cite{haddad2019wind} accommodating
non-stationary and multivariate data. However, these statistical models are limited in their ability to capture complex patterns and long-range dependencies. Neural network models such as RNN \cite{wang2020day} and LSTM \cite{jahangir2020deep} introduced feedback-driven structures to handle sequential data in order to get around these restrictions. By using gating mechanisms to address the vanishing gradient issue, LSTMs improved the preservation of long-term dependencies.
However, due to their dependence on sequential processing, RNNs and LSTMs both struggle with computational inefficiency. Models based on convolutional neural networks (CNNs), including Temporal Convolutional Networks (TCN) \cite{lara2020temporal}, TimesNET \cite{zuo2023ensemble}, and WaveNet \cite{wang2023improved}, used convolutions to leverage parallel processing and provide efficiency gains. Nevertheless, they continued to face difficulties while modeling long-range dependencies.

In order to address these issues, Transformers \cite{vaswani2017attention} implemented attention mechanisms that effectively capture intricate temporal dependencies. PatchTST \cite{nie2022time}, enhanced forecasting accuracy by employing patch-based techniques and performing cross-domain learning. Autoformer \cite{wu2021autoformer} combined time series decomposition with attention. iTransformer \cite{liu2023itransformer} improved temporal pattern modeling by using adaptive sparse attention and learning dynamic interaction across time series. Transformers have shown their abilities in predicting different power system time series, such as load \cite{xu2023power}, renewable generation \cite{wu2022interpretable}, and electricity prices \cite{azam2021multi}. However, their scalability in applications that require long-range modeling is limited by their model size and the quadratic complexity $O(n^2)$ of attention mechanisms. Because of these limitations, there has been a growing interest in linear models that provide strong performance while maintaining computational efficiency. Models like TiDE \cite{das2023long}, DLinear \cite{zeng2023transformers}, and NLinear \cite{zeng2023transformers} achieved a balance between predictive accuracy and scalability by combining deep learning with linear decomposition techniques.

Mamba \cite{gu2023mamba} and deep state space models (SSMs) \cite{gu2021efficiently} have recently emerged as efficient alternatives to Transformers, surpassing them in a variety of tasks. These models enhance scalability and efficiency with parallel training and sub-quadratic inference \cite{wang2024mamba}. In order to manage dependencies at various scales, the State Space Transformer (SST) model \cite{xu2024integrating} integrates Mamba with Transformers, while SiMBA \cite{patro2024simba} employs Fast Fourier Transform (FFT) for channel modeling. Bi-Mamba+ \cite{liang2024bi} introduces dynamic adaptation and bidirectional processing for intra- and inter-series correlations, and TimeMachine \cite{timemachine} employs four Mamba blocks with varying resolutions, which enables efficient multivariate forecasting.

Building on these developments, our model offers three significant innovations that improve multivariate prediction with Mamba-based models. First, it uses a dynamic time series decomposition to separate seasonal and trend components and capture the underlying dynamics of highly variable time series.
Second, fixed-size linear projections of the input and output sequences ensure that our model size does not scale with increasing context or prediction window sizes.
Third, the model employs \textbf{\textit{two parallel}} Mamba blocks, one of which is inversed to process transposed time series. This inverse operation enables dual tokenization, where one captures intra-series dependencies within each time series, while the other captures inter-series relationships across time slots.

As forecasting methods have advanced, the need for comprehensive datasets that match the capabilities of modern machine learning models has grown, but existing datasets often have significant limitations in the types of time series they cover and their duration. For instance, \cite{zheng2022multi} offers one year of data for load and renewable energy generation in ERCOT and MISO, while \cite{building900k} provides synthetic data spanning one year for 900,000 buildings that is designed for short-term load forecasting. In \cite{STLF}, six years of day-ahead electricity prices are provided in separate datasets for markets such as France, Germany, and PJM, where  \cite{pombo2022benchmarking} focuses on a single solar plant in Denmark, offering one year of data for short-term PV-power forecasting. While these datasets provide valuable insight, they do not have the granularity, synchronization, and multivariate scope needed for advanced machine learning models. Our dataset resolves these issues by integrating zonal loads, zonal electricity prices, ancillary service prices, and wind and solar generation into one dataset that spans five years with hourly resolution. This dataset is published along with our toolbox and together they provide a platform to benchmark and foster innovation in power system forecasting. While this work focuses on zonal medium- to long-term forecasting tasks typically performed by Transmission System Operators (TSOs), the proposed model is general and can be adapted for short-term forecasting and for applications at the Distribution System Operator (DSO) level as well.

\section{Mamba and State Space Models}
\label{Mamba}
SSMs are a class of sequence modeling frameworks inspired by classical state space representations of continuous systems. They are powerful tools in power systems, offering a flexible framework to model dynamic behaviors, capture temporal dependencies under uncertainty, and enable real-time estimation of complex system states \cite{kundur1994stability,5667072, 4539776}. Kalman filters, a widely used application of SSMs, have been extensively applied for dynamic state estimation \cite{abur2004state}. Autoregressive SSMs also play a critical role in time series forecasting in power systems \cite{contreras2003arima}. Many physical processes in the grid, including load evolution and renewable dynamics, can be described by differential or difference equations, which SSMs are naturally equipped to represent. This makes SSMs particularly well-suited for modeling latent drivers of grid time series that are not directly observable but evolve based on structured interactions and external inputs. Building on this foundation, this work introduces deep SSMs as a powerful extension to capture the underlying dynamics and interdependencies of multivariate time series in power systems.

\subsection{State Space Models}
SSMs use first-order differential equations to transform an input function \( x(t) \in \mathbb{R}^D \) into an output function \( y(t) \in \mathbb{R}^D \) via an intermediate latent state \( h(t) \in \mathbb{R}^N \). The governing equations in the continuous domain are:
\begin{equation}
\frac{\mathrm{d}h(t)}{\mathrm{d}t} = Ah(t) + Bx(t), \quad y(t) = Ch(t),
\end{equation}
where \( A \in \mathbb{R}^{N \times N} \), \( B \in \mathbb{R}^{N \times D} \), and \( C \in \mathbb{R}^{D \times N} \) are learnable parameters that define the system's dynamics. To enable discrete-time modeling, the continuous parameters \( A \) and \( B \) are discretized into \( \bar{A} \) and \( \bar{B} \) using a zero-order hold (ZOH) method with a sampling interval \( \Delta \):
\begin{equation}
\bar{A} = \exp(\Delta A), \quad \bar{B} = A^{-1}(\exp(\Delta A) - I)B,
\end{equation}
where \( I \) represents the identity matrix and \( \exp(\cdot) \) denotes the matrix exponential. In the discrete-time domain, the state update and output equations are reformulated as:
\begin{equation}
h_t = \bar{A} h_{t-1} + \bar{B} x_t, \quad y_t = Ch_t,
\end{equation}
where \( h_t \) is the latent state at time step \( t \). This discretization turns the continuous-time modeling to a more practical implementation for sequence modeling tasks. Training this discretized recurrent form of the SSM is challenging due to its sequential dependency, where the computation of \( h_t \) depends directly on \( h_{t-1} \). To overcome this limitation, we exploit the linear time-invariant property of the system and reformulate the continuous convolution into a discrete one. For simplicity, we initialize the system state with \( x_{-1} = 0 \). By unrolling the dynamics, the hidden states and corresponding outputs of the system are expressed as:
\begin{align}
h_0 &= Bx_0, \quad & y_0 &= C h_0, \nonumber\\
h_1 &= ABx_0 + Bx_1, \quad & y_1 &= C h_1, \nonumber\\
h_2 &= A^2Bx_0 + ABx_1 + Bx_2, \quad & y_2 &= C h_2.
\end{align}
We can compactly represent this process as a convolutional operation. In particular, the output \( y_t \) can be expressed as:
\begin{eqnarray}
y_t = CA^tBx_0 + CA^{t-1}Bx_1 + \cdots + CABx_{t-1} + CBx_t,
\end{eqnarray}
or equivalently $y = K \ast x$, where \( K \) represents the convolution kernel, defined as:
\begin{eqnarray}
K \in \mathbb{R}^L := \big(CA^iB\big)_{i=0}^{L-1} = \big(CB, CAB, \ldots, CA^{L-1}B\big). 
\end{eqnarray}
This formulation reduces the problem to a single convolution operation, which enables parallelized training and efficient inference analogous to that of deep neural networks. This convolution-based representation also allows SSMs to process an input token \(\mathbf{x} \in \mathbb{R}^{L \times D}\) with linear complexity \(O(LD)\), while Transformers scale quadratically \(O(L^2D)\) because their self-attention computes pairwise interactions across all \(L \times L\) tokens, making SSMs significantly more efficient for long sequences. The model's trainable parameters \( \bar{A} \), \( \bar{B} \), \( C \), and \( \Delta \) are learned from data. Prior research demonstrates that initializing \( \bar{A} \) with a special matrix called the HIPPO matrix significantly enhances the model's ability to capture long-term dependencies, thereby introducing a new variant of SSM known as the structured state space model (S4)\cite{gu2021efficiently}.

\subsection{Mamba}

SSMs employ static matrices $\bar{A}$, $\bar{B}$, and $\bar{C}$ that are fixed after training, and do not change based on the input context. Although this static design is effective for certain tasks, it restricts their capacity to dynamically adjust focus based on the significance of input. For example, if we use a context window of one week to predict a time series, all seven days are treated equally, despite the fact that days with higher volatility have a greater predictive value. A static formulation lacks flexibility to prioritize these critical patterns while deemphasizing less relevant information. The Mamba model, which was introduced in \cite{gu2023mamba}, resolves this limitation by transforming $B$, $C$, and the discretization interval $\Delta$ into input-dependent functions.  As a result, $\bar{A}$ becomes input-dependent as well. In particular, the parameters $B$ and $C$ are updated for an input token $\mathbf{x}$ using the function $B, C \leftarrow \text{LinearN}(\mathbf{x})$, where $\text{LinearN}$ is a learnable N-dimensional linear projection function. Similarly, the discretization interval $\Delta$ adjusts to the input by employing a softplus transformation.
\begin{equation}
\Delta = \text{softplus}(\text{parameter} + \text{LinearD}(\text{Linear1}(\mathbf{x}))).
\end{equation}
This design is the basis of selective SSMs and Mamba, which are capable of dynamically adapting to the input and focusing on the more important parts while forgetting the less relevant data. In the next section, we introduce our prediction model, which improves on the Mamba model to better capture time series patterns.

\section{PowerMamba Architecture}
\label{PowerMamba}
\begin{figure}   
\vspace{-2mm}
\begin{center}
\includegraphics[width=.75\columnwidth]{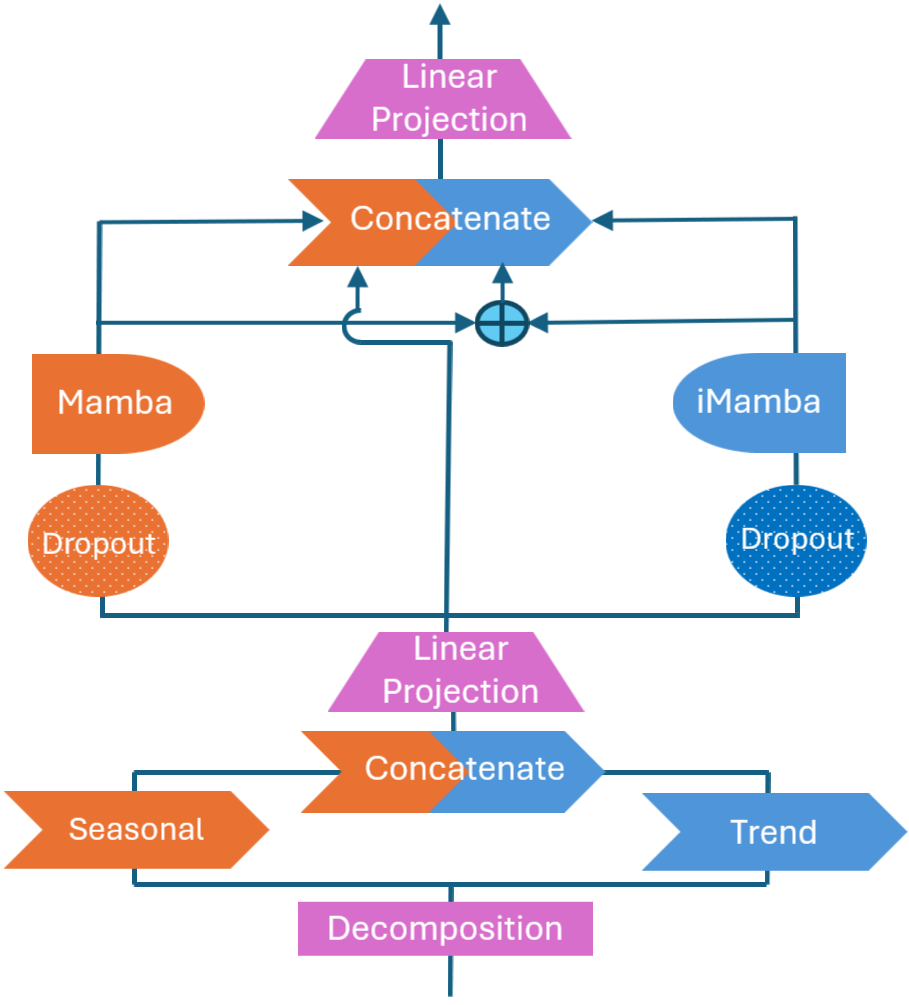}
\par\end{center}
\vspace{-4mm}
\caption{PowerMamba architecture. This model uses decompositions along with linear projections, and a combination of normal and inverse Mamba blocks to improve prediction.}
\vspace{-5mm}
\label{fig:architecture}
\end{figure}

Let us consider each point in time as \(x_t \in \mathbb{R}^D\), where \(D\) denotes the number of channels in the multivariate time series. Using a context or look-back window 
\(\mathbf{x} = [x_1, x_2, \dots, x_L] \in \mathbb{R}^{L \times D}\), our goal is to predict 
\([x_{L+1}, x_{L+2}, \dots, x_{L+W}] \in \mathbb{R}^{W \times D}\), where \(L\) and \(W\) are the look-back and prediction window sizes, repectively. In this setting, we are only using the historical records in the context window to perform the prediction, but later in Section~\ref{ext_pred}, we extend our model to incorporate external predictions as well. In the following, we describe the architecture of our prediction model named PowerMamba, which is shown in Figure~\ref{fig:architecture}.

\subsection{Data Normalization}
We first standardize the time series data before feeding it into our model. The Z-score normalization is usually performed on the the data to have a unit variance and zero mean. However, it might be difficult for the Z-score normalization to handle the dynamic shifts in the distribution of the time series data. In order to resolve this issue, we implement Reversible Instance Normalization (RevIN) \cite{kim2022reversible}, which is more suitable for time series with fluctuating statistical properties. RevIN modifies each data point by considering each feature at different time points and is defined as:
\begin{equation}
x_{t}^d = \gamma_d \left( \frac{x_{t}^d - \text{mean}(x_{t}^d)}{\sqrt{\text{var}(x_{t}^d) + \epsilon}} \right) + \beta_d,
\end{equation}
where \(\gamma_d\) and \(\beta_d\) are learnable parameters of channel \(d\) that allow RevIN to tailor the normalization to the unique characteristics of the dataset. Our model can perform both types of normalization, but our findings indicate that RevIN outperforms Z-score normalization.

\subsection{Decomposition and Projection}
\label{decomposition}
To decompose a time series, we apply a moving average kernel to the input sequence and obtain a trend component \( T(\mathbf{x}) = \text{Moving Average}(\mathbf{x}) \). Furthermore, we calculate a seasonal component, indicated as \( S(\mathbf{x}) = \mathbf{x} - T(\mathbf{x}) \), which quantifies the difference between the original sequence and the trend component. The moving average window size plays a critical role in the quality of the decomposition, where smaller windows may capture noise, while larger ones risk oversmoothing important patterns. We treat the window size as a hyperparameter and select it by performing a grid search to optimize forecasting accuracy. Following the time series decomposition, we combine these two inputs to create a longer input, \(\mathbf{x}_{ts} = [T(\mathbf{x}); S(\mathbf{x})] \), which is \(2L \times D\) in size. Next, we use a linear projection $\mathbf{x}_e = \text{LinearE}(\mathbf{x}_{ts})$
to project the input into an embedding of size \(E \times D\) before entering the mamba block. The fixed size of this embedding ensures that the length of the vector entering the mamba block does not depend on the input size. Hence, the size of the Mamba block does not change for different input sizes.
The value of the embedding E is a model hyperparameter that can be found with grid search.


\subsection{The Mamba Blocks}
Two parallel Mamba blocks are included in our model, where the left Mamba block receives inputs of size \( L \times D \), and the right Mamba block uses an inverse Mamba (iMamba) to process the inputs of size \( D \times L \), which are the transposed versions of the input. A dropout layer is placed before each of these Mamba blocks to prevent overfitting. With the combination of these two types of blocks, we create content-dependent features within each channel and between different channels. We formulate this process as follows:
\begin{eqnarray}
\mathbf{x}_{m} = \textit{Mamba} ({DO}_1(\mathbf{x}_e)), \, \, \, \,\mathbf{x}_{im} = \textit{iMamba} ({DO}_2(\mathbf{x}_e^T)),
\end{eqnarray} 
where \( \mathbf{x}_{m} \) and \( \mathbf{x}_{im} \) represent the outputs of regular and inverse Mamba blocks, \( DO \) denotes the dropout operation, and \( \mathbf{x}_e^T \) is the transposed embedded input. The input dimensions are preserved by both blocks, guaranteeing that their outputs correspond to the original size. To match the input dimensions, the iMamba block output is transposed back. This architecture effectively generates and processes \textit{temporal tokens}, representing multivariate information for each time step, and \textit{variate tokens}, which embed the entire time series of each variate independently into a token. By combining these tokenizations, the model captures channel correlations and local context while simultaneously learning long-range global dependencies.

\subsection{Output Projection}
After receiving the output tokens from the Mamba blocks, our objective is to project these tokens to generate predictions of size \( W \times D \). In addition to the outputs of the two Mamba blocks, we incorporate an element-wise addition of these outputs, denoted as \( \mathbf{x}_{m} \bigoplus \mathbf{x}_{im} \), to enhance the model's representation by integrating complementary features from both Mamba blocks. This operation further improves gradient flow and contributes to the training stability of the model. As illustrated in Figure \ref{fig:architecture}, we also employ a residual connection to retain the original input information (\( \mathbf{x}_e \)) before it enters the blocks. Together with the outputs of the Mamba blocks, these two tokens are concatenated to produce an extensive vector that contains all the required information needed to do the final prediction:
\begin{equation}
    \mathbf{x}_{c} = [\mathbf{x}_e; \mathbf{x}_{m}; \mathbf{x}_{im}; \mathbf{x}_{m} \oplus \mathbf{x}_{im}]
\end{equation}
The concatenated vector \( \mathbf{x}_{c} \) is then used to predict the final output \( \mathbf{y} \) through a linear projection $\mathbf{y} = \textit{LinearW}(\mathbf{x}_{c})$. Using linear projections both before and after the Mamba blocks, combined with a fixed embedding for the Mamba models, ensures that the model size does not scale with an increase in either the context length or the prediction window size.

\begin{figure}   
\begin{center}
\includegraphics[width=.6\columnwidth]{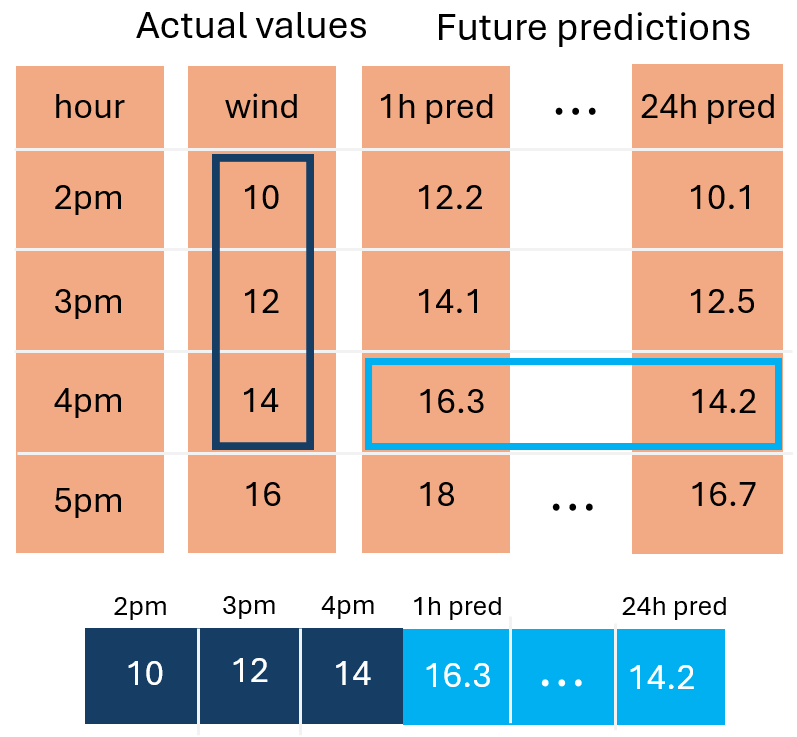}
\par\end{center}
\vspace{-3mm}
\caption{\label{fig:pred_dataset} The structure of the dataset with external predictions. For a context window ending at 4 PM, the corresponding 24-hour external predictions are extracted.}
\vspace{-6mm}
\end{figure}

\subsection{Integrating External Predictions}
\label{ext_pred}

Consider a multivariate time series dataset with \( D \) channels, each containing only historical data. Including external predictions for \( W \) future hours increases the dataset size by \( W \) additional columns per channel, resulting in a total increase from \( D \) to \( (W+1)D \). For instance, as shown in Figure \ref{fig:pred_dataset}, at 4 PM the
dataset includes actual values for the past hours and the external predictions of
the next 24 hours. This results in a 25-fold increase in size for \( W = 24 \). This direct integration is computationally inefficient, so we
propose a framework to restructure and compress input tokens, avoiding redundancies while preserving relevant information.

\textbf{Appending Predictions to the Sequence}: For each token with its specific context window, instead of including all \( W \) prediction columns for each channel, we extract only the predictions relevant to that context. Specifically, for a context window ending at time \( t \), predictions for \( t+1, t+2, \ldots, t+W \) are appended directly to the context window as shown in Figure \ref{fig:pred_dataset}. This transformation increases the temporal dimension from \( L \) to \( L+W \) while maintaining the number of channels at \( D \). It also eliminates redundancy and guarantees that external predictions are temporally in line with the context.

\textbf{Incorporating Predictions into Context Features}: In most prediction models, multivariate tokens are generated by aggregating data from all channels for a specific time step. In such a setting, the external predictions cannot be included in the tokens as distinct features since we have only appended them to the end of their corresponding context sequence. To resolve
this issue, for each channel, we create an additional column where we shift the external predictions back into the context window. This process enables including future predictions into the context window as additional features to help improve prediction. We define this transformation as:
\begin{eqnarray}
&&x_{\textit{exp}}[:, :D] = x_{\textit{app}}, \,\, \quad x_{\textit{exp}}[:, D:2D] = x_{\textit{app}}, \quad and \nonumber\\
&&x_{\textit{exp}}[1:L, D:2D] = x_{\textit{app}}[W:L+W, D:2D] .
\end{eqnarray}
Here, \( x_{\textit{app}} \) is the input token with appended predictions and size \((L+W) \times D \), and \( x_{\textit{exp}} \) is the expanded dataset with size \((L+W) \times 2D \), ensuring predictions for future steps are included in the input tokens.

\textbf{Compressing Tokens to Fixed Length}: Finally, in order to prevent the token size from expanding excessively, the expanded input's length is projected back to the original context length \( L \), while the enriched predictive information is retained. After this stage, the input token size is reduced to \(L \times 2D \), which makes it independent of \( W \). This keeps model complexity manageable, even for long prediction horizons, while maintaining accurate forecasting information.

The unique innovations introduced in our model, including external forecast integration, cross-channel representation through the inverse Mamba block, and trend-seasonal decomposition, make it particularly well suited for power system forecasting. These elements enable the model to effectively handle exogenous variability, dynamic correlations, and nonstationary patterns across grid variables.

\section{Comprehensive Electric Grid Dataset}
\label{dataset}
In this section, we introduce GridSet, which is a five-year multivariate dataset of the ERCOT electric grid. It covers 43,824 hours (1,826 days) from 2019
to 2023, and it includes zonal loads (8 columns), zonal electricity prices (8 columns), ancillary service prices (4 columns), and renewable generation (wind and solar, 2
columns) with an hourly resolution. We use the first four years of the dataset for
training, and reserve the the final year for testing. This timeframe captures the growth of renewable generation, energy storage, and large flexible loads in Texas,
reflecting their increasing impact on the grid. Each time series displays distinct dynamics, as seen in Figure \ref{data_plot} in the Appendix. Load patterns follow weekly and daily cycles, whereas wind and solar generation are extremely unpredictable, posing unique modeling challenges. Table \ref{tab:datasets} summarizes the main statistical
characteristics of the dataset. Here, we provide an overview of the four groups of time series that constitute this dataset.

\subsection{Zonal Loads}
The ERCOT region has eight different weather zones, and the load is reported for each of these zones. Their power consumption could be categorized into three main sectors: industrial, commercial, and residential. Given the disparities in weather patterns and the uneven distribution of population and industrial activities within Texas, load patterns exhibit substantial variations among the zones. Therefore, in order to capture their distinct consumption patterns, it is imperative
to analyze individual regions. 



\begin{table}[t]
\centering
\caption{Overview of the characteristics of our dataset.}
\resizebox{\linewidth}{!}{
\begin{tabular}{l|c|c|c}
\toprule
Time Series& Channels&Mean value &Standard deviation \\
\midrule
Zonal loads (MW)& 8 & 5797.93 & 1446.52 \\
Zonal electricity prices (\$/MWh)& 8 & 67.06 & 422.15 \\
Ancillary service prices (\$/MWh)  & 4 & 52.41 & 662.53\\
Renewable generation (MW)& 2 & 6363.30 & 4438.36 \\
\bottomrule
\end{tabular}
}
\label{tab:datasets}
\vspace{-2mm}
\end{table}


\begin{table*}[t]
\centering
\caption{Results in MSE and MAE (the lower, the better) using the comprehensive grid dataset. The results are compared with baseline models under different prediction window lengths ($W = {24, 48, 72, 96, 168}$). The input sequence length ($L$) is fixed at 240 for all baselines. The best results are highlighted in bold, and the second-best results are \underline{underlined}.}
\vspace{-3mm}
\label{tab:result}
\setlength{\tabcolsep}{2pt}
\medskip
\resizebox{.85\linewidth}{!}{
\begin{tabular}{lc|cc|cc|cc|cc|cc|cc|cc|cc|cc|cc|cc|cc} 
\toprule
          
\multicolumn{2}{c}{Methods$\rightarrow$}&\multicolumn{2}{c|}{PowerMamba} & \multicolumn{2}{c|}{TimeMachine} & \multicolumn{2}{c|}{iTransformer} & \multicolumn{2}{c|}{PatchTST} & \multicolumn{2}{c|}{DLinear} & \multicolumn{2}{c|}{TimesNET}  & \multicolumn{2}{c|}{Autoformer} {}\\ 
\midrule
$\mathcal{D}$& $T$ & MSE & MAE & MSE & MAE & MSE & MAE & MSE & MAE & MSE & MAE & MSE & MAE & MSE & MAE \\
\midrule
\multirow{5}{*}{\rotatebox{90}{GridSet}}
&24&\bf0.129&\bf0.166&\underline{0.135}&\underline{0.166}&0.147&{0.181}&0.142&0.183&0.168&0.199&0.175&0.198& 0.301&0.311\\
&48&\bf0.174&\bf0.199&\underline{0.181}&\underline{0.200}&0.193&0.213&0.202&0.225&0.230&0.245&0.220&0.231& 0.326&0.335\\
&72&\bf0.201&\bf0.220&\underline{0.207}&\underline{0.220}&0.219&0.232&0.235&0.248&0.294&0.287&0.252&0.247& 0.329&0.330\\
&96&\bf0.219&\bf0.232&\underline{0.233}&\underline{0.238}&{0.233}&0.240&0.244&0.252&0.338&0.310&0.259&0.253& 0.327&0.332\\
&168&\bf0.242&\underline{0.247}&\underline{0.247}&\bf0.246&0.251&0.253&0.253&0.259&0.385&0.348&0.274&0.261& 0.326&0.330\\
\midrule
\multirow{5}{*}{\rotatebox{90}{Load}}
&24&\bf0.098&\bf0.223&\underline{0.102}&\underline{0.227}&0.108&0.237&0.113&0.240&0.134&0.268&0.136&0.275& 0.238&0.365\\
&48&\bf0.152&\bf0.280&\underline{0.158}&\underline{0.284}&0.161&0.291&0.174&0.301&0.228&0.358&0.198&0.333& 0.330&0.438\\
&72&\bf0.193&\bf0.318&0.202&\underline{0.325}&\underline{0.200}&0.326&0.215&0.337&0.368&0.453&0.249&0.368& 0.358&0.454\\
&96&\bf0.222&\bf0.343&0.235&0.352&\underline{0.225}&\underline{0.348}&0.235&0.353&0.474&0.505&0.270&0.385& 0.357&0.454\\
&168&\bf0.265&\bf0.378&0.272&\underline{0.378}&\underline{0.265}&0.381&{0.269}&0.383&0.588&0.579&0.307&0.412& 0.371&0.461\\
\midrule
\multirow{5}{*}{\rotatebox{90}{Price}}
&24&\bf0.100&\underline{0.082}&0.108&\bf0.079&\underline{0.107}&0.090&0.109&0.102&0.120&0.101&0.137&0.090& 0.215&0.194\\
&48&\bf0.128&\underline{0.095}&\underline{0.131}&\bf0.089&0.133&0.101&0.154&0.124&0.143&0.112&0.157&0.105& 0.192&0.180\\
&72&\bf0.146&{0.107}&0.148&\bf0.096&\underline{0.146}&0.107&0.184&0.138&0.154&0.118&0.173&\underline{0.104}& 0.195&0.169\\
&96&\underline{0.154}&{0.108}&0.164&{0.106}&\bf0.152&\underline{0.106}&0.190&0.133&0.161&0.122&0.164&\bf0.103& 0.194&0.175\\
&168&\underline{0.165}&0.109&0.167&{0.104}&\bf0.155&\underline{0.104}&0.174&0.123&0.166&0.137&0.180&\bf0.102& 0.189&0.168\\
\midrule
\multirow{5}{*}{\rotatebox{90}{AS Price}}
&24&\bf0.019&\bf0.029&\underline{0.020}&\underline{0.029}&0.021&{0.033}&0.022&0.038&0.022&0.037&0.028&0.032& 0.089&0.153\\
&48&\bf0.024&\underline{0.035}&\underline{0.025}&\bf0.032&0.026&0.036&0.032&0.047&0.027&0.044&0.029&0.036& 0.058&0.129\\
&72&\bf0.028&\underline{0.038}&\underline{0.028}&\bf0.034&0.029&0.039&0.040&0.055&0.030&0.049&0.035&0.040& 0.049&0.109\\
&96&\bf0.030&0.041&0.031&\underline{0.038}&\underline{0.030}&\bf0.038&0.041&0.055&{0.031}&0.053&0.033&0.039& 0.048&0.113\\
&168&\underline{0.032}&0.040&0.033&\underline{0.038}&\bf0.030&\bf0.037&0.037&0.048&0.034&0.071&0.033&0.034& 0.042&0.102\\
\midrule
\multirow{5}{*}{\rotatebox{90}{Renewables}}
&24&\bf0.590&\bf0.545&\underline{0.610}&\underline{0.550}&0.712&0.620&0.623&0.570&0.783&0.638&0.786&0.659& 1.317&0.878\\
&48&\bf0.750&\bf0.621&\underline{0.778}&\underline{0.637}&0.891&0.701&0.847&0.684&0.994&0.730&0.944&0.717& 1.383&0.951\\
&72&\bf0.799&\bf0.649&\underline{0.821}&\underline{0.661}&0.968&0.741&0.906&0.718&1.087&0.788&1.013&0.747& 1.315&0.918\\
&96&\bf0.841&\bf0.667&\underline{0.904}&\underline{0.705}&0.992&0.750&0.899&0.713&1.124&0.800&1.044&0.758&  1.293&0.907\\
&168&\bf0.883&\bf0.685&\underline{0.899}&\underline{0.698}&1.019&0.767&0.943&0.733&1.143&0.822&0.997&0.442& 1.264&0.904\\
\bottomrule
\end{tabular}
} \vspace{-5mm}
\end{table*}

\subsection{Zonal Electricity Prices}
Our dataset includes day-ahead settlement point prices for eight different weather zones. Settlement point prices are the weighted average of the LMPs across different nodes within
an ERCOT zone, and for the remainder of this paper, we refer to them as electricity prices. Each ERCOT region has its own unique characteristics that distinguish it from other regions in terms of electricity price patterns. For instance, the "West" zone experiences numerous negative price hours as a result of the high wind production during the night, whereas the "Houston" zone experiences frequent price spikes as a result of the high residential and industrial power consumption and the increasing limitations in the transmission capacity.

\subsection{Ancillary Service Prices}
Our dataset includes four ancillary service time series corresponding to the following products: Regulation Up (Reg-Up), Regulation Down (Reg-Down), Responsive Reserve Service (RRS), and Non-Spinning Reserve Service (Non-Spin). These ancillary services are procured in the day-ahead market
with an hourly resolution. Reg-Up and Reg-Down provide frequency regulation, which addresses minor supply-demand mismatches within the 5-minute Economic Dispatch intervals.
In contrast, RRS and Non-Spin act as emergency reserves, and they are only 
activated during significant reserve shortfalls \cite{menati2023modeling}.

\subsection{Renewable Generation}
Hourly solar and wind generation for the ERCOT region is also included in the dataset. These time series are particularly difficult to predict because of their complex patterns and their substantial reliance on weather conditions. As it can be seen in Table \ref{tab:datasets}, the standard deviation of renewables is relatively large
compared to its mean value, indicating substantial variations over time. Additionally, the sharp increase in wind and solar integration within ERCOT has altered the long-term patterns of these time series. This variability across different years in the dataset captures valuable patterns, which contribute to training more robust models capable of adapting to the evolving nature
of renewable energy generation.

\subsection{External Predictions}
To enhance the accuracy of energy management, ERCOT provides load and renewable generation forecasts developed using advanced models that integrate climate simulations and weather predictions. These forecasts are updated hourly, offering predictions with increasing accuracy as the target hour approaches, and extend up to one week ahead with an hourly resolution. In this work, we include ERCOT's zonal load predictions (8 columns) and renewable generation forecasts for wind and solar (2 columns). However, no external forecasts are available for electricity or ancillary service prices. Our dataset utilizes a 24-hour window of these external predictions, although the framework is designed to support any configurable prediction window size. Incorporating these forecasts expands the dataset from $22$ features to $22 + 10 \times 24 = 262$ features.

\begin{table*}[t]
\centering
\caption{Comparison of the prediction accuracy of our model and the baselines with and without external predictions. A fixed context size of $L=240$ and a prediction window size of $W=24$ are used for all the baselines.} \label{tab:result_prediction} 
\vspace{-1mm}
\setlength{\tabcolsep}{2pt} \medskip \resizebox{.85\linewidth}{!}
{ \begin{tabular}{lc|cc|cc|cc|cc|cc|cc|cc|cc|cc|cc|cc|cc} \toprule \multicolumn{2}{c}{Methods$\rightarrow$}&\multicolumn{2}{c|}{PowerMamba} & \multicolumn{2}{c|}{TimeMachine} & \multicolumn{2}{c|}{iTransformer} & \multicolumn{2}{c|}{PatchTST} & \multicolumn{2}{c|}{DLinear} & \multicolumn{2}{c|}{TimesNET} & \multicolumn{2}{c|}{Autoformer} {}\\ \midrule $\mathcal{D}$& PR& MSE & MAE & MSE & MAE & MSE & MAE & MSE & MAE & MSE & MAE & MSE & MAE & MSE & MAE  \\ \midrule \multirow{1}{*}{{GridSet}} &\xmark&\bf0.129&\bf0.166&\underline{0.135}&\underline{0.166}&0.147&0.181&0.142&0.183&0.168&0.199&0.175&0.198&0.301&0.311\\  \multirow{1}{*}{{GridSet}} &\cmark&\bf0.074&\bf0.123&\underline{0.080}&\underline{0.126}&0.091&0.144&0.109&0.157&0.113&0.166&0.109&0.148 &0.113&0.170\\ 
\midrule \multirow{1}{*}{{Load}} &\xmark&\bf0.098&\bf0.223&\underline{0.102}&\underline{0.227}&0.108&0.237&0.113&0.240&0.134&0.268&0.136&0.275 &0.238&0.365\\  \multirow{1}{*}{{Load}} &\cmark &\bf0.065&\bf0.183&\underline{0.071}&\underline{0.188}&0.091&0.215&0.123&0.244&0.114&0.240&0.086&0.215 &0.090&0.219\\ 
\midrule \multirow{1}{*}{{Price}}&\xmark &\bf0.100&\underline{0.082}&0.108&\bf0.079&\underline{0.107}&0.090&0.109&0.102&0.120&0.101& 0.137&0.090 &0.215&0.194\\  \multirow{1}{*}{{Price}} &\cmark&\bf0.093&\bf0.075&0.099&\underline{0.075}&\underline{0.096}&0.082&0.107&0.085&0.113&0.095&0.122&0.079 &0.133&0.117\\ 
\midrule \multirow{1}{*}{{ASPrice}} &\xmark&\bf0.019&\bf0.029&\underline{0.020}&\underline{0.029}&0.021&0.033&0.022&0.038&0.022&0.037&0.028&0.032  &0.089&0.153\\  \multirow{1}{*}{{ASPrice}} &\cmark&\bf0.018&\underline{0.029}&\underline{0.018}&\bf{0.026}&0.018&0.030&0.019&0.029&0.020&0.038&0.028&0.032 &0.033&0.076\\ 
\midrule \multirow{1}{*}{{Renewables}} &\xmark&\bf0.590&\bf0.545&\underline{0.610}&\underline{0.545}&0.712&0.620&0.623&0.570&0.783&0.638&0.786&0.659 &1.317&0.878\\  \multirow{1}{*}{{Renewables}} &\cmark&\bf0.142&\bf0.267&\underline{0.162}&\underline{0.279}&0.218&0.332&0.241&0.353&0.299&0.406&0.314&0.392&0.282&0.373\\ 
\bottomrule
\end{tabular}
}
\vspace{-2mm}
\end{table*}

\section{Experiments}
\label{results}
\subsection{Experiment Setup}
We demonstrate the advanced features of our open-access prediction toolbox and perform our analysis using Mean squared error (MSE) and mean absolute error (MAE) as evaluation metrics. These metrics are based on standardized time series to make a fair comparison by removing the impact of scales and magnitudes. Our analysis is conducted in two main scenarios: one that excludes external predictions from the dataset and another that incorporates these
predictions. The performance of PowerMamba is compared with a variety of baseline models, each of which is constructed using a distinct architectural approach. These baselines include TimeMachine \cite{timemachine}, which is based on the Mamba architecture, iTransformer \cite{liu2023itransformer} and PatchTST \cite{nie2022time}, which are Transformer-based models, Autoformer \cite{wu2021autoformer}, which uses linear transformers, TimesNet \cite{zuo2023ensemble}, which relies on temporal convolutional networks (TCNs), and DLinear \cite{zeng2023transformers}, a linear neural network-based model. We implemented all the using the Time Series Library\footnote{https://github.com/thuml/Time-Series-Library/} and PyTorch. In order to have a fair comparison, the same grid search was conducted for all baselines, and all the models were trained using an L2 loss function and the ADAM optimization algorithm \cite{kingma2014adam}. The batch size changed for each model, but the training process was consistently conducted for 50 epochs. Additional experiments are presented in the Appendix, including a comprehensive evaluation on the PJM ISO, training and inference time comparisons, robustness to external forecast errors, and ablation studies on the impact of model size and the decomposition module.


\begin{figure}[htbp]
    \centering
    \includegraphics[width=\columnwidth]{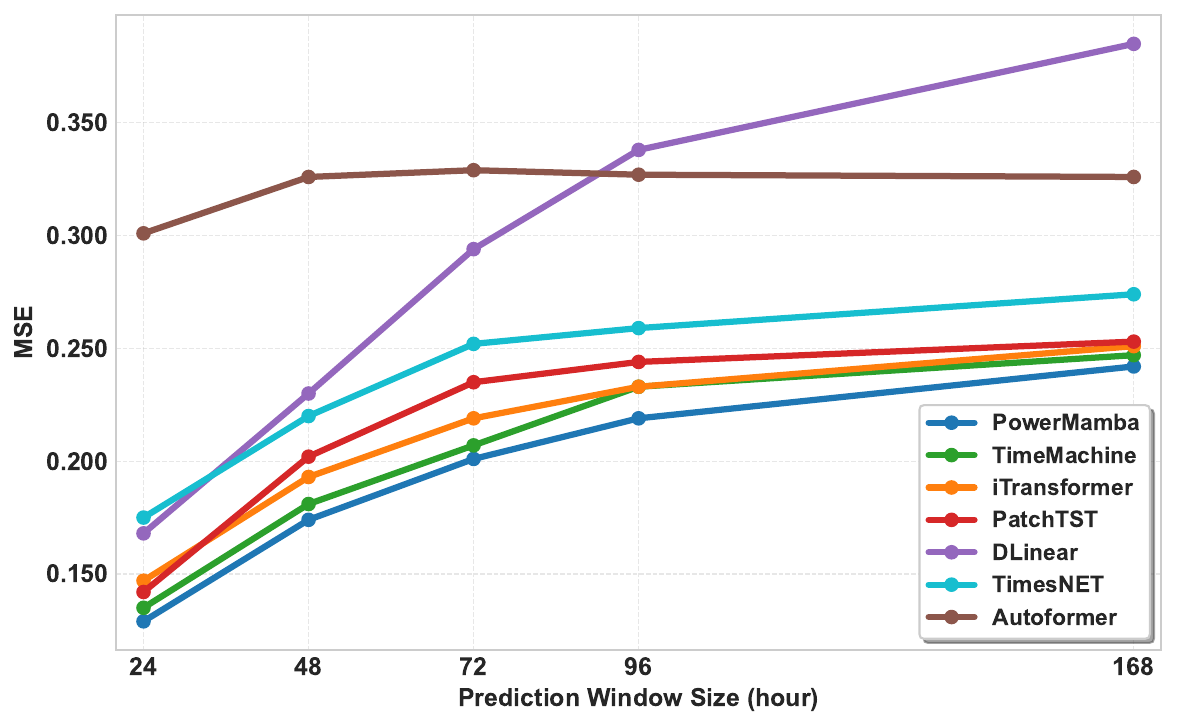}
    \vspace{-3mm}
    \caption{MSE of all models for different prediction window sizes, with context length $L=240$ and no external predictions.}
    \label{fig:prediction_window}
    \vspace{-2mm}
\end{figure}

\subsection{Main Results Without External Predictions}
In this section, external predictors are excluded from the dataset, and the model utilizes only the data available within the look-back window. The comprehensive forecasting results
are summarized in Table \ref{tab:result}, with the best-performing models (lowest MSE and MAE) highlighted in bold and the second-best underlined. We report the accuracy of each model for GridSet and its four distinct categories by calculating the average of the time series within them. According to the results, our model consistently outperforms all other models across the entire dataset. Furthermore, Mamba-based models exhibit a clear performance advantage compared to non-Mamba models. Our model is particularly effective at predicting renewable generation and loads, which are notoriously difficult to predict due to their inherent variability. In terms of electricity price and ancillary service price predictions, our model outperforms other models for prediction windows of 72 hours or shorter and is either the best or second-best for longer windows. Given the wide range of dynamics and different statistical characteristics of these four categories, our model’s high accuracy across all of them demonstrates its adaptability and robustness.

\begin{figure}[htbp]
    \centering
    \includegraphics[width=.96\columnwidth]{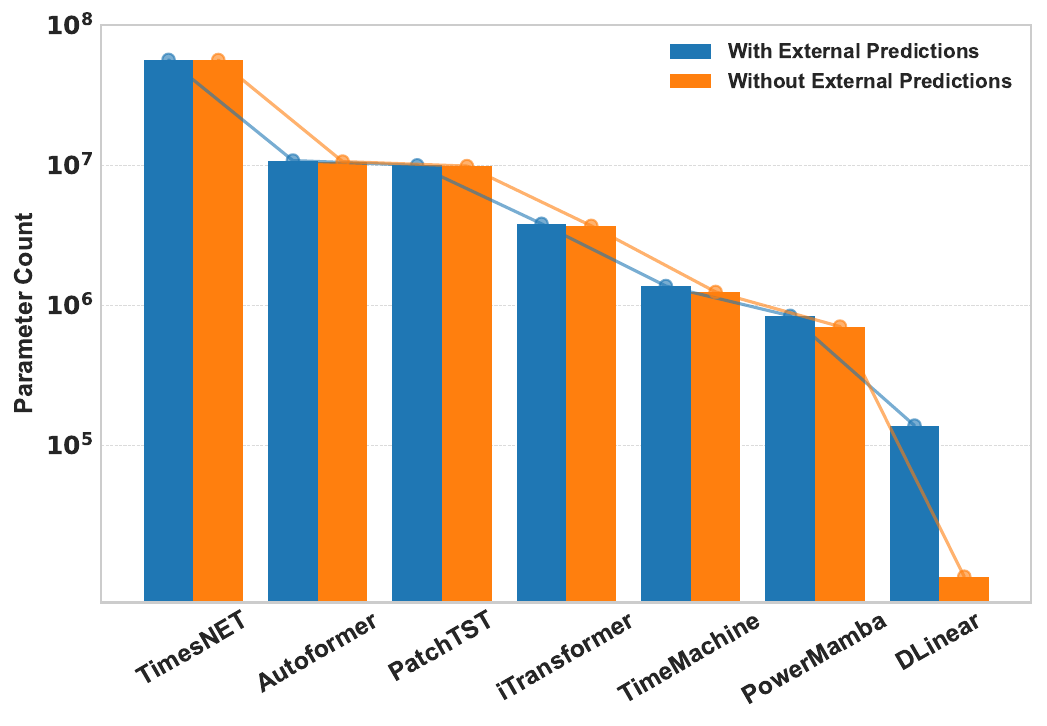}
    \vspace{-3mm}
    \caption{The number of model parameters with and without external predictions, in log scale.}
    \vspace{-2mm}
    \label{fig:param_count}
\end{figure}

The prediction accuracy of all models is depicted in Figure \ref{fig:prediction_window} for a variety of prediction window sizes. PowerMamba consistently outperforms other models in both short- and long-term predictions. As demonstrated, the prediction error increases as the window size increases, a trend that is consistent across all models. The performance of the majority of models decreases at a similar rate, with two exceptions. DLinear, a linear model that functions well for shorter prediction windows but experiences a rapid decline in accuracy for longer horizons, and Autoformer, a linear Transformer-based model that consistently underperforms across all window sizes.

\subsection{Main Results With External Predictions}

In this section, we integrate the external predictions for load and renewable generation into our dataset. Our external prediction processing block is added to each baseline model to process input tokens before they enter the models. In Table \ref{tab:result_prediction}, we compare the prediction accuracy of our model and the baselines for a 24-hour prediction window, both with and without external predictions. The results indicate that our model performs better than all baselines in all four categories and the entire dataset by incorporating external predictions, resulting in a 43\% reduction in its overall prediction error. All time series categories, including those without external predictions, are affected by this enhancement. For instance, the renewable generation prediction error is reduced by 76\%, and price prediction error is reduced by 7\%, despite the absence of external price predictions. These findings demonstrate how time series within the electric grid are correlated and how well our multivariate predictions capture the interactions between different factors.

\begin{figure*}[htbp]
    \centering
    \begin{subfigure}[b]{0.45\textwidth}
        \centering
        \includegraphics[width=\textwidth]{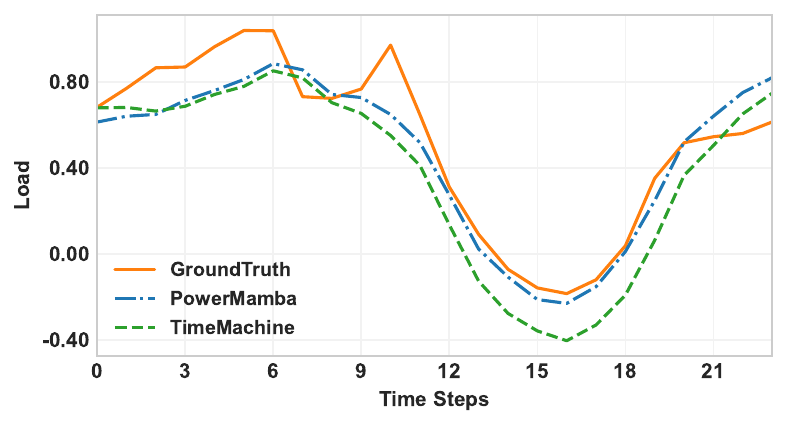}
        \caption{Load}
        \label{fig:all_models}
    \end{subfigure}%
    \hfill
    \begin{subfigure}[b]{0.45\textwidth}
        \centering
        \includegraphics[width=\textwidth]{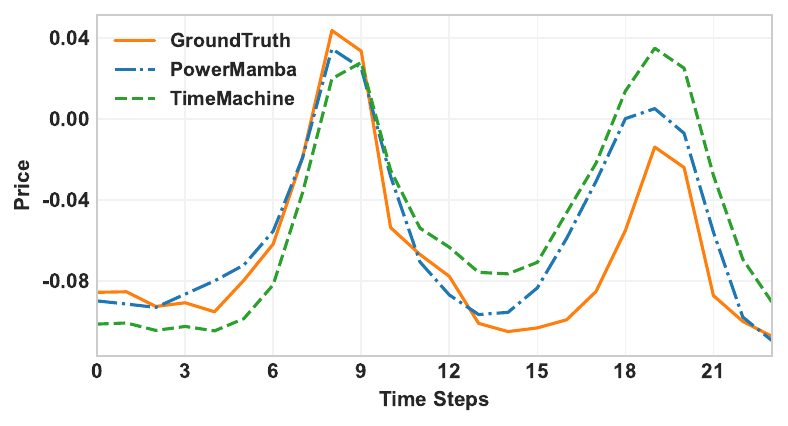}
        \caption{Price}
        \label{fig:smallest_models}
    \end{subfigure}
    \vspace{-1mm}
    \caption{The ground truth and 24-hour predictions of load and price by PowerMamba and TimeMachine for a fixed context size of $L=240$. Our model effectively captures the trend and provides predictions closest to the ground truth.}
    \label{fig:2_channel}
    \vspace{-2mm}
\end{figure*}

\begin{figure*}[htbp]
    \centering
        \begin{subfigure}[b]{0.45\textwidth}
        \centering
        \includegraphics[width=\textwidth]{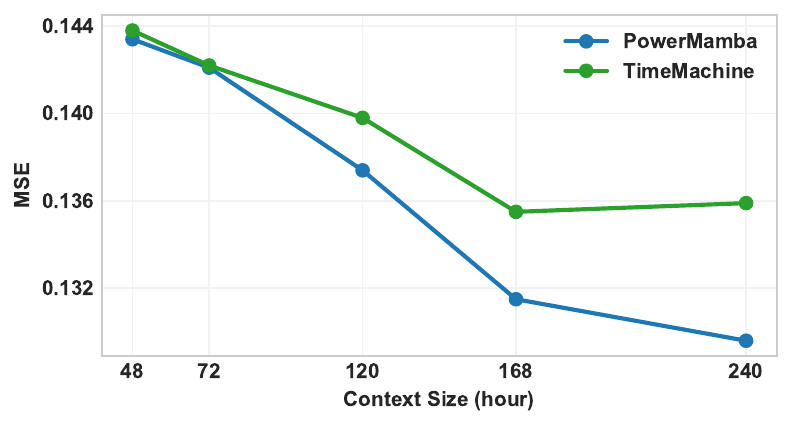}
        \caption{Without external predictions}
        \label{fig:without_prediction}
    \end{subfigure}%
    \hfill
    \begin{subfigure}[b]{0.45\textwidth}
        \centering
        \includegraphics[width=\textwidth]{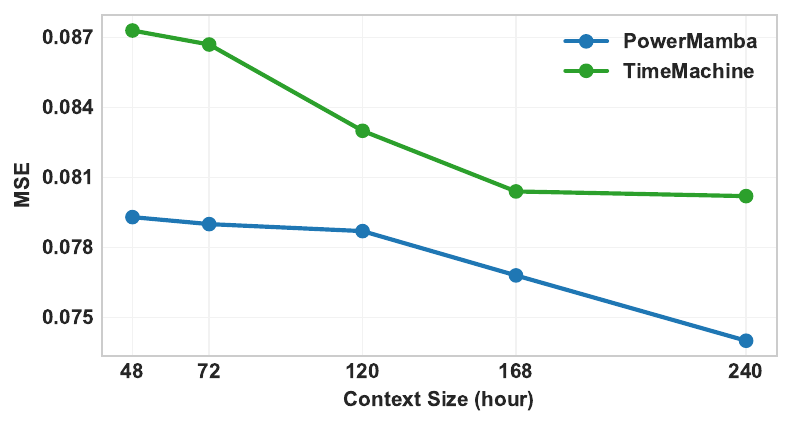}
        \caption{With external predictions}
        \label{fig:with_prediction}
    \end{subfigure}
    \caption{The impact of context size on the MSE of PowerMamba and TimeMachine (the second-best performing model) for a 24-hour prediction window. The accuracy of our models consistently improves with longer context sizes.}
    \label{fig:context_size}
    \vspace{-2mm}
\end{figure*}

\subsection{Additional Experiments}

\textbf{Computational Efficiency:}
Given the increasing computational demands of advanced machine learning models, it is necessary to evaluate their architectural efficiency by counting the number of trainable parameters changed by gradient descent algorithms during training. In this section, the number of trainable parameters are compared for all the baselines, both with and without external predictions. The context window size for all the models is fixed at 240 hours, while the prediction
window size is 24 hours. Figure \ref{fig:param_count} displays the results in log scale, emphasizing the substantial disparities in efficiency. 

Among all competitive models, PowerMamba is the smallest, with DLinear being the only smaller baseline, though it is a
linear model and unable to compete with larger models. PowerMamba enhances the prediction accuracy of TimeMachine by 7\% while employing 43\% fewer parameters, thereby highlighting its superiority over this state-of-the-art Mamba-based model. Furthermore, PowerMamba is considerably smaller than Transformer-based models, with 78\% fewer parameters than the leading iTransformer. Additionally, including our external prediction processing block into different
models results in minimal changes to their size, demonstrating the block’s efficiency in improving prediction capabilities of sequence-to-sequence models.

\textbf{Qualitative Comparison:} 
Figure \ref{fig:2_channel} presents a visual comparison of a randomly selected 24-hour prediction window for load and price that has been generated by PowerMamba and TimeMachine. Although both models are able to accurately depict load and price patterns, PowerMamba exhibits superior alignment, especially around peaks and troughs. PowerMamba performs better in price prediction, where there are usually larger deviations from the ground truth, by capturing more of the peak variations and remaining closer to the actual values. On the other hand, TimeMachine captures fewer variations and experiences difficulties in following the ground truth, particularly at critical points.

 
\textbf{Impact of context window size:} In this part, we investigate the influence of varying context window sizes to determine the extent to which increased contextual information affects model performance and to assess the models' ability to effectively utilize this information. As shown in Figure \ref{fig:context_size}, the MSE of both models decreases by increasing the context window size, which demonstrates the effectiveness of Mamba-based architecture's in managing long-range dependencies. However, PowerMamba shows a greater ability to use extended context by consistently improving as the look-back window increases. Additional experiments and supporting analyses on multiple aspects of the model are provided in the Appendix.

\section{Conclusion}
\label{conclusion}
This paper introduces PowerMamba, a Deep State Space Model for multivariate time series prediction in power systems. The model is substantially smaller than other competitive benchmarks in terms of the number of parameters and can integrate external predictions without increasing its size. Our results show that PowerMamba outperforms state-of-the-art prediction models across a variety of power-system prediction tasks. To support further research, we release an open-access toolbox along with a comprehensive, high-resolution time series dataset to benchmark a wide range of advanced machine learning models. For future work, we plan to expand the dataset to include additional time series from different resources, such as energy storage systems and data centers. Additionally, future work could investigate the use of deep state-space models to capture transient power-system dynamics and improve the prediction of dynamic disturbances.

\section*{Disclaimer}
The views expressed in this paper are the opinion of the
authors and do not reflect the views of PJM Interconnection,
L.L.C. or its Board of Managers of which Le Xie is a member.

{
\appendix

\subsection{Experiments with Public Data from PJM}

In this section, we extend our analysis beyond ERCOT and construct a large-scale forecasting dataset for the PJM ISO, incorporating renewable generation (wind and solar), zonal loads, and zonal day-ahead electricity prices over six years from 2019 to 2024. Due to space constraints, we exclude external forecasts from the PJM experiments and focus solely on historical input data. For the PJM experiments, we conducted a fresh grid search for PowerMamba and all baseline models to ensure optimal hyperparameter settings under this new system configuration. The comprehensive forecasting results are summarized in Table~\ref{tab:pjm_result}. As shown, PowerMamba consistently outperforms all baseline models across various time series and prediction window sizes. Figure~\ref{fig:prediction_window_pjm} illustrates the MSE trends for different prediction lengths, further confirming PowerMamba's superior performance in both short- and long-term forecasting tasks. These results are important because PJM represents a fundamentally different grid environment from ERCOT, with a distinct market structure, fuel mix, and load behavior. Evaluating PowerMamba across both ERCOT and PJM, two of the largest ISOs in the United States, highlights the model’s strong generalization capabilities. Furthermore, each ISO contains zones with highly heterogeneous characteristics. By assessing performance across this spatial diversity, we demonstrate the model’s robustness in capturing complex dynamics under a wide range of operating conditions.

\begin{table}[t]
\centering
\caption{Prediction results for the PJM dataset across various tasks and prediction window lengths ($W = \{24, 48, 72, 96, 168\}$) with context length ($L$) fixed at 240.}
\vspace{-3mm}
\label{tab:pjm_result}
\setlength{\tabcolsep}{2pt}
\medskip
\resizebox{\linewidth}{!}{
\begin{tabular}{lc|cc|cc|cc|cc} 
\toprule
\multicolumn{2}{c}{Methods$\rightarrow$} & \multicolumn{2}{c|}{PowerMamba} & \multicolumn{2}{c|}{TimeMachine} & \multicolumn{2}{c|}{iTransformer} & \multicolumn{2}{c}{PatchTST} \\
\midrule
$\mathcal{D}$ & $T$ & MSE & MAE & MSE & MAE & MSE & MAE & MSE & MAE \\
\midrule
\multirow{5}{*}{\rotatebox{90}{PJM dataset}} 
& 24  & \textbf{0.141} & \textbf{0.223} & \underline{0.149} & \underline{0.228} & 0.168 & 0.241 & 0.159 & 0.239 \\
& 48  & \textbf{0.214} & \textbf{0.280} & \underline{0.223} & \underline{0.284} & 0.259 & 0.304 & 0.234 & 0.297 \\
& 72  &  \textbf{0.262} & \textbf{0.313} & \underline{0.273} & \underline{0.316} & 0.309 & 0.338 & 0.293 & 0.338 \\
& 96  & \textbf{0.296} & \textbf{0.333} & \underline{0.319} & \underline{0.344} & 0.347 & 0.359 & 0.339 & 0.370 \\
& 168 & \textbf{0.361} & \textbf{0.369} & \underline{0.375} & \underline{0.376} & 0.403 & 0.393 & 0.383 & 0.394 \\
\midrule
\multirow{5}{*}{\rotatebox{90}{Load}} 
& 24  & \textbf{0.100} & \textbf{0.218} & \underline{0.108} & \underline{0.225} & 0.121 & 0.240 & 0.124 & 0.242 \\
& 48  & \textbf{0.171} & \textbf{0.288} & \underline{0.174} & \underline{0.291} & 0.201 & 0.312 & 0.199 & 0.311 \\
& 72  & \textbf{0.218} & \textbf{0.330} & \underline{0.225} & \underline{0.333} & 0.253 & 0.356 & 0.252 & 0.353 \\
& 96  & \textbf{0.255} & \textbf{0.358} & \underline{0.272} & \underline{0.368} & 0.292 & 0.384 & 0.292 & 0.382 \\
& 168 & \textbf{0.333} & \textbf{0.410} & \underline{0.338} & \underline{0.413} & 0.360 & 0.430 & 0.347 & 0.421 \\
\midrule
\multirow{5}{*}{\rotatebox{90}{Price}} 
& 24  & \textbf{0.122} & \textbf{0.193} & \underline{0.128} & \underline{0.196} & 0.141 & 0.203 & {0.136} & 0.204 \\
& 48  & \textbf{0.186} & \textbf{0.236} & \underline{0.198} & \underline{0.240} & 0.220 & 0.251 & 0.200 & 0.250 \\
& 72  & \textbf{0.233} & \textbf{0.262} & \underline{0.243} & \underline{0.262} & 0.265 & 0.277 & 0.263 & 0.290 \\
& 96  & \textbf{0.261} & \textbf{0.273} & \underline{0.290} & \underline{0.286} & 0.298 & 0.291 & 0.314 & 0.326 \\
& 168 & \textbf{0.317} & \textbf{0.297} & \underline{0.337} & \underline{0.307} & 0.340 & 0.314 & 0.347 & 0.344 \\
\midrule
\multirow{5}{*}{\rotatebox{90}{Renewables}} 
& 24  & \textbf{0.786} &  \underline{0.614} & 0.817 & 0.625 & 0.981 & 0.703 &  \underline{0.787} & \textbf{0.613} \\
& 48  &  \underline{0.990} &  \underline{0.700} & 1.023 & 0.713 & 1.313 & 0.822 & \textbf{0.980} & \textbf{0.694} \\
& 72  & \textbf{1.054} & \textbf{0.722} & 1.115 & 0.747 & 1.420 & 0.861 &  \underline{1.065} &  \underline{0.732} \\
& 96  &  \underline{1.127} & \textbf{0.749} & 1.148 & 0.760 & 1.491 & 0.885 & \textbf{1.113} &  \underline{0.752} \\
& 168 & \textbf{1.165} & \textbf{0.763} & 1.193 & 0.777 & 1.587 & 0.919 &  \underline{1.181} &  \underline{0.779} \\
\bottomrule
\end{tabular}
}
\end{table}

\begin{figure}[htbp]
    \centering
    \vspace{-3mm}
    \includegraphics[width=\columnwidth]{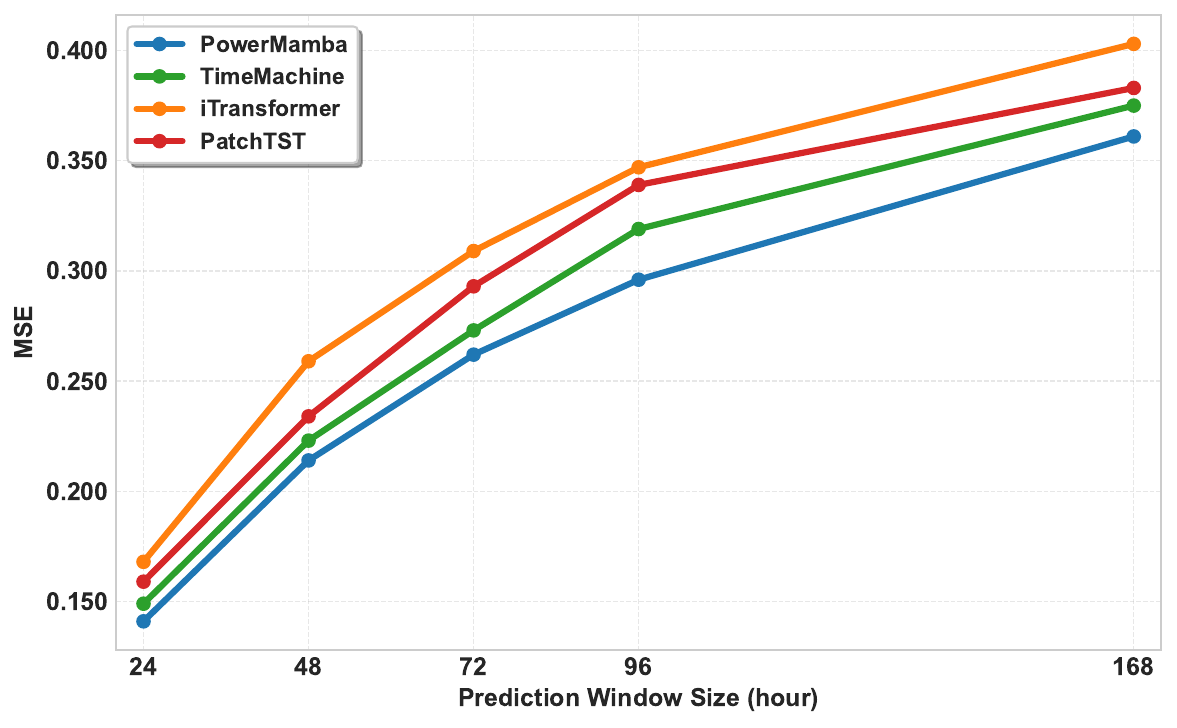}
    \vspace{-6mm}
    \caption{MSE of all models for different prediction window sizes, with context length $L=240$ and no external predictions.}
    \label{fig:prediction_window_pjm}
    \vspace{-6mm}
\end{figure}

\subsection{Training and Inference Time}
Figure~\ref{fig:train_loss_comparison} shows the training loss convergence curves for both variants of PowerMamba with and without external forecasts. The results show smooth and consistent reductions in training loss over epochs, indicating stable optimization behavior across runs. To verify the computational efficiency of PowerMamba, we measured the total training time and per-instance inference latency across all baseline models. PowerMamba achieves 19\% faster training and 14\% faster inference compared to the best Mamba-based baseline (TimeMachine), and over 60\% and 57\% improvements respectively compared to the fastest transformer-based model (iTransformer). While DLinear achieves the fastest training and inference times overall, it is a lightweight linear model that cannot match the performance of larger architectures. PowerMamba’s competitive model size contributes to lower training costs and faster inference, making it a strong candidate for real-time deployment in power system operations.

\begin{figure}[htbp]
\vspace{-3mm}
    \centering
    \includegraphics[width=\linewidth]{fig/training_loss_comparison.pdf}
    \vspace{-6mm}
    \caption{Training loss convergence for PowerMamba with and without external predictions.}
    \label{fig:train_loss_comparison}
    \vspace{-1mm}
\end{figure}

\subsection{Effect of External Prediction Errors}
To evaluate our model’s robustness to uncertainty in external prediction inputs, we simulate realistic prediction noise during inference. Specifically, Gaussian noise is injected into all input columns that have an external prediction available, such as zonal load, wind, and solar generation. The noise is applied after scaling but before inference, without retraining the model, to reflect real-world scenarios where models are trained on clean data but operate with imperfect external forecasts. The perturbation is applied multiplicatively as:
\[
\tilde{x} = x + \mathcal{N}(0, \sigma \cdot |x|),
\]
with noise levels $\sigma$ representing relative forecast error magnitudes. PowerMamba maintains strong prediction accuracy across a wide range of noise intensities, as shown in Figure~\ref{fig:mse_noise}. The model shows only a 2.6\% increase in MSE with 10\% forecast noise and remains stable even with 20\% noise. It is important to note that in our setup, noise is injected independently at each time step within the external forecast, resulting in highly erratic inputs. In contrast, real-world ensemble forecasts tend to exhibit coherent temporal patterns that more closely resemble the true trajectory. As such, injecting 30\% noise into our external forecast presents a significantly more challenging scenario than typical ensemble variability. Yet, even under this condition, PowerMamba remains stable, with MSE increasing by no more than 25.9\%, highlighting model’s robustness to uncertainty in external forecasts.

\begin{table}[t]
\centering
\caption{Training and inference time for different models with $W = 24$ on the ERCOT dataset.}
\label{tab:efficiency_comparison}
\vspace{-1mm}
\scriptsize
\resizebox{0.5\textwidth}{!}{
\begin{tabular}{lcc}
\toprule
\textbf{Model} & \textbf{Total Train Time (min)} & \textbf{Inference Time / Instance (ms)} \\
\midrule
PowerMamba   & 11.56   & \(0.113\) \\
TimeMachine  & 14.22   & \(0.131\) \\
iTransformer & 28.65   & \(0.260\) \\
PatchTST     & 55.98   & \(0.781\) \\
Autoformer   & 134.43  & \(1.84\) \\
TimesNET     & 1773.03 & \(13.9\) \\
DLinear      & 8.98    & \(0.020\) \\
\bottomrule
\end{tabular}
}
\vspace{-4mm}
\end{table}

\begin{figure}[htbp]
\vspace{-3mm}
    \centering
    \includegraphics[width=\linewidth]{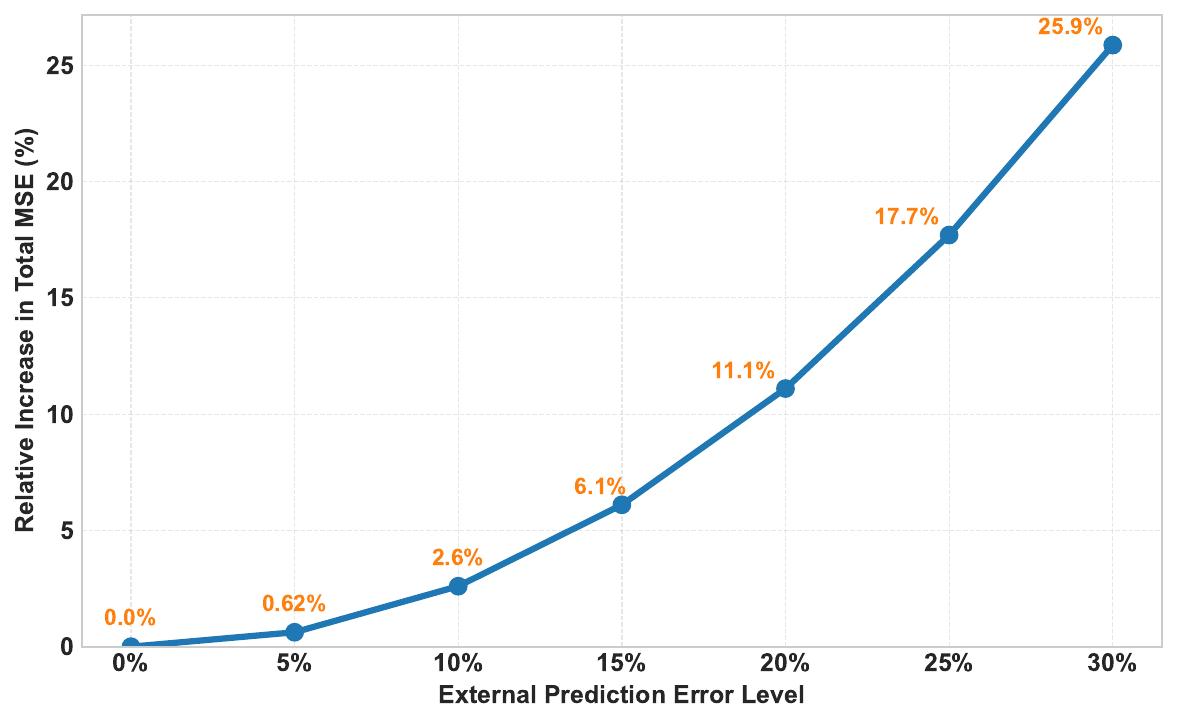}
    \vspace{-6mm}
    \caption{Percentage change in the MSE of PowerMamba for different levels of external prediction error.}
    \label{fig:mse_noise}
    \vspace{-1mm}
\end{figure}

\begin{table}[htbp]
\centering
\caption{The Impact of model size on performance with $W = 96$ and no external predictions.}
\vspace{-1mm}
\label{tab:embed_param_mse_reduced}
\small
\resizebox{\linewidth}{!}{%
\begin{tabular}{c|ccc}
\toprule
\textbf{\# Parameters} 
& \textbf{MSE} & \textbf{TrendMSE ($\sigma{=}6$)} & \textbf{TrendMSE ($\sigma{=}24$)} \\
\midrule
121k  & 0.233 & 0.134 & 0.093 \\
248k  & 0.225 & 0.130 & 0.089 \\
581k  & 0.225 & 0.130 & 0.090 \\
1.5M & 0.230 & 0.135 & 0.093 \\
3M    & 0.248 & 0.149 & 0.107 \\
\bottomrule
\end{tabular}}
\vspace{-4mm}
\end{table}

\subsection{The Impact of Model Size}
In this section, we analyze how forecasting performance changes by adjusting the size of the PowerMamba model, which is measured by the number of trainable parameters. Specifically, we vary the internal embedding size of the Mamba blocks from 64 to 512, resulting in total parameter counts ranging from 120K to 3 million. The number of parameters significantly impacts training efficiency, memory usage, and deployment cost.  All models are trained on the entire ERCOT dataset without external forecasts and evaluated at 96-hour prediction window sizes. As it can be seen in Table \ref{tab:embed_param_mse_reduced}, performance improves as model size increases, but begins to plateau or slightly degrade beyond 1M parameters. This slight decline is likely due to overfitting, as larger models require significantly more training data to generalize effectively. 

To investigate the impact of model size on PowerMamba’s ability in modeling long-term trends, we apply Gaussian filtering to both the predicted and ground-truth sequences. This technique smooths the data by replacing each value with a weighted average of its neighbors, where the weights follow a Gaussian distribution. The degree of smoothing is controlled by the kernel width $\sigma$, where smaller values such as 6 hours preserve moderate fluctuations, but larger values such as 24 hours emphasize persistent low-frequency patterns. By applying this filtering, we isolate trend dynamics and assess the model’s ability to capture long-term trends independent of short-term noise. 

We report both standard MSE and trend-focused MSE (TrendMSE) across two smoothing levels. The trend-focused MSE is measured by comparing the trend component of the ground truth with that of the predicted time series. Results indicate that PowerMamba captures long-term patterns effectively at all model sizes. Importantly, smaller variants with lower than 500k parameters perform on par with larger models, showing that the architecture efficiently models low-frequency behavior without requiring large parameter counts. These findings support PowerMamba’s ability in capturing long-range trends, which is essential for power system applications.

\subsection{Impact of the Decomposition Module}
The decomposition module divides the time series into trend and seasonal components, making it easier to handle noise and short-term variations. As can be seen in Figure \ref{data_plot}, each time series in our dataset exhibits very distinct dynamics, making it difficult to model their variations without decomposition. To measure the impact of the decomposition module in the performance of PowerMamba, we will examine how the model handles highly variable inputs. We define an anomaly score for each input sequence $x_i \in \mathbb{R}^{L \times D}$ as: 
\[
\delta_i = \max_{l,d} \left| \frac{x_{i,l,d} - \mu_d}{\sigma_d} \right|,
\]
where $\mu_d$ and $\sigma_d$ denote the mean and standard deviation of variable $d$ across the dataset. We select the top 5\% of input instance with the highest $\delta_i$ scores to represent extreme and highly variable inputs.
We compare two model variants, one with decomposition and one without, and compute the MSE across these samples. As shown in Table~\ref{tab:extreme_input_mse}, the model with decomposition consistently outperforms the baseline, particularly for volatile time series. The MSE improves by more than 6 percent for renewables and around 5 percent for prices. As we expect, the improvement is negligible for zonal loads, which are generally more smooth time series. These findings confirm that the decomposition module improves model robustness to sudden fluctuations and enhances prediction accuracy under real-world conditions.

\begin{table}[t]
\centering
\caption{MSE of PowerMamba with and without  decomposition on the top 5\% of highly variable input samples.}
\label{tab:extreme_input_mse}
\vspace{-1mm}
\scriptsize
\resizebox{0.5\textwidth}{!}{
\begin{tabular}{lccc}
\toprule
\textbf{Metric (MSE)} & \textbf{No Decomp} & \textbf{With Decomp} & \textbf{\% Improvement} \\
\midrule
Price (Top 5\%)     & 0.483 & 0.459 & 4.97 \\
Load (Top 5\%)      & 0.095 & 0.095 & 0.09 \\
Renewable (Top 5\%) & 0.399 & 0.375 & 6.01 \\
Total (Top 5\%)     & 0.276 & 0.262 & 5.07 \\
\bottomrule
\end{tabular}
}
\vspace{-4mm}
\end{table}

\begin{figure}   
\begin{center}
\includegraphics[width=\linewidth]{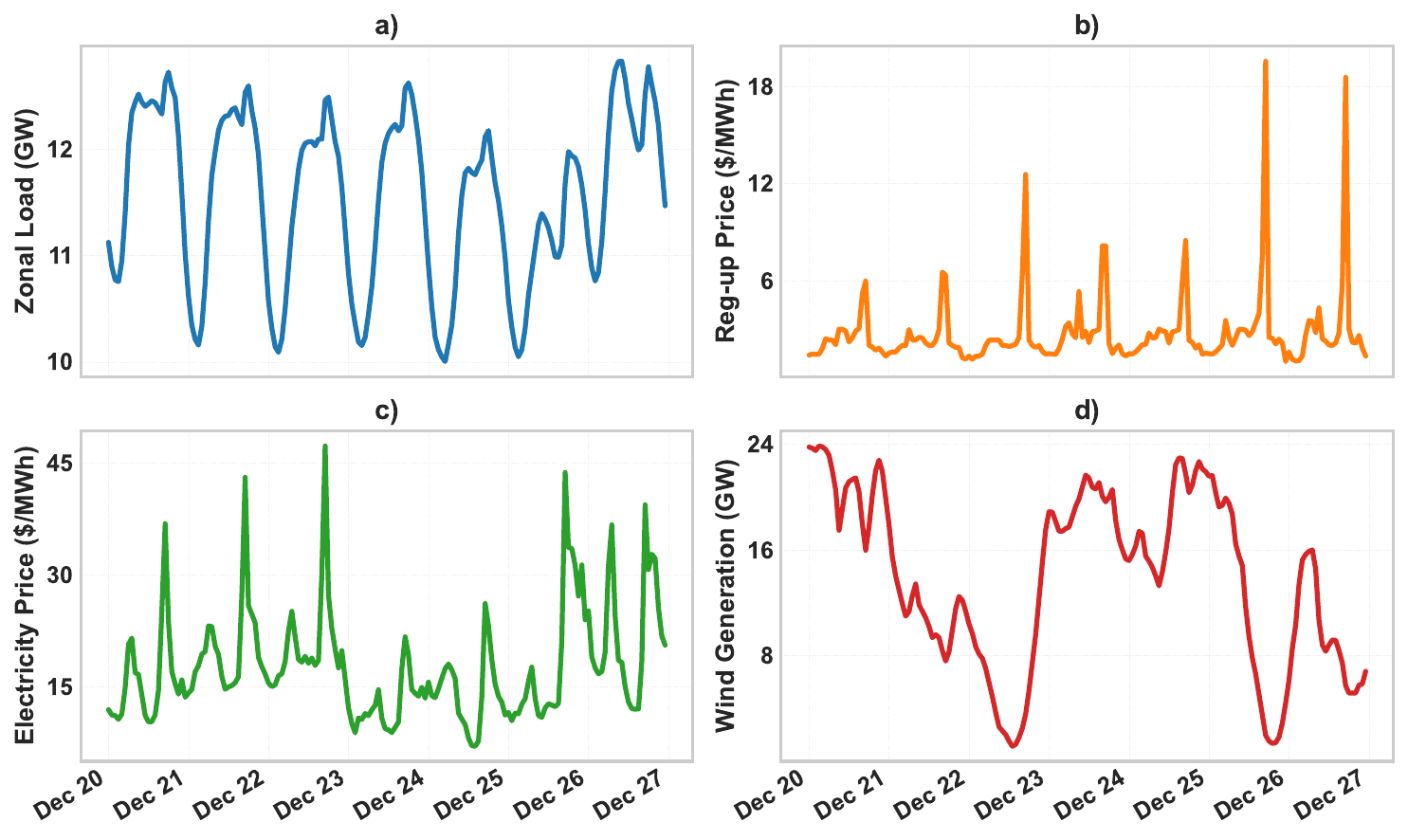}
\end{center}
\vspace{-1mm}
\caption{An example of the four types of time series for the Houston zone during a week in December 2023. (a) Load, (b) Reg-up price, (c) Electricity price, (d) Wind generation. \label{data_plot}}
\vspace{-4mm}
\end{figure}

}

\Urlmuskip=0mu plus 1mu\relax  
\bibliographystyle{IEEEtran}
\bibliography{ref}

@article{liu2023itransformer,
  title={itransformer: Inverted transformers are effective for time series forecasting},
  author={Liu, Yong and Hu, Tengge and Zhang, Haoran and Wu, Haixu and Wang, Shiyu and Ma, Lintao and Long, Mingsheng},
  journal={arXiv preprint arXiv:2310.06625},
  year={2023}
}

@article{nie2022time,
  title={A time series is worth 64 words: Long-term forecasting with transformers},
  author={Nie, Yuqi and Nguyen, Nam H and Sinthong, Phanwadee and Kalagnanam, Jayant},
  journal={arXiv preprint arXiv:2211.14730},
  year={2022}
}

@article{wu2021autoformer,
  title={Autoformer: Decomposition transformers with auto-correlation for long-term series forecasting},
  author={Wu, Haixu and Xu, Jiehui and Wang, Jianmin and Long, Mingsheng},
  journal={Advances in neural information processing systems},
  volume={34},
  pages={22419--22430},
  year={2021}
}

@article{das2023long,
  title={Long-term forecasting with tide: Time-series dense encoder},
  author={Das, Abhimanyu and Kong, Weihao and Leach, Andrew and Mathur, Shaan and Sen, Rajat and Yu, Rose},
  journal={arXiv preprint arXiv:2304.08424},
  year={2023}
}

@inproceedings{zeng2023transformers,
  title={Are transformers effective for time series forecasting?},
  author={Zeng, Ailing and Chen, Muxi and Zhang, Lei and Xu, Qiang},
  booktitle={Proceedings of the AAAI conference on artificial intelligence},
  volume={37},
  number={9},
  pages={11121--11128},
  year={2023}
}

@article{wang2024mamba,
  title={Is mamba effective for time series forecasting?},
  author={Wang, Zihan and Kong, Fanheng and Feng, Shi and Wang, Ming and Yang, Xiaocui and Zhao, Han and Wang, Daling and Zhang, Yifei},
  journal={arXiv preprint arXiv:2403.11144},
  year={2024}
}

@article{patro2024simba,
  title={Simba: Simplified mamba-based architecture for vision and multivariate time series},
  author={Patro, Badri N and Agneeswaran, Vijay S},
  journal={arXiv preprint arXiv:2403.15360},
  year={2024}
}

@article{xu2024integrating,
  title={Integrating Mamba and Transformer for Long-Short Range Time Series Forecasting},
  author={Xu, Xiongxiao and Liang, Yueqing and Huang, Baixiang and Lan, Zhiling and Shu, Kai},
  journal={arXiv preprint arXiv:2404.14757},
  year={2024}
}

@article{liang2024bi,
  title={Bi-Mamba4TS: Bidirectional Mamba for Time Series Forecasting},
  author={Liang, Aobo and Jiang, Xingguo and Sun, Yan and Lu, Chang},
  journal={arXiv preprint arXiv:2404.15772},
  year={2024}
}

@article{STLF,
   title={Forecasting day-ahead electricity prices: A review of state-of-the-art algorithms, best practices and an open-access benchmark},
   volume={293},
   ISSN={0306-2619},
   DOI={10.1016/j.apenergy.2021.116983},
   journal={Applied Energy},
   publisher={Elsevier BV},
   author={Lago, Jesus and Marcjasz, Grzegorz and De Schutter, Bart and Weron, Rafał},
   year={2021},
   month=jul, pages={116983} }

@inproceedings{building900k,
author = {Emami, Patrick and Sahu, Abhijeet and Graf, Peter},
title = {BuildingsBench: a large-scale dataset of 900K buildings and benchmark for short-term load forecasting},
year = {2024},
publisher = {Curran Associates Inc.},
address = {Red Hook, NY, USA},
booktitle = {Proceedings of the 37th International Conference on Neural Information Processing Systems},
articleno = {871},
numpages = {35},
location = {New Orleans, LA, USA},
series = {NIPS '23}
}

@inproceedings{patchtst,
title={A Time Series is Worth 64 Words:  Long-term Forecasting with Transformers},
author={Yuqi Nie and Nam H Nguyen and Phanwadee Sinthong and Jayant Kalagnanam},
booktitle={International Conference on Learning Representations },
year={2023},
url={https://openreview.net/forum?id=Jbdc0vTOcol}
}

@article{kingma2014adam,
  title={Adam: A method for stochastic optimization},
  author={Kingma, Diederik P and Ba, Jimmy},
  journal={arXiv preprint arXiv:1412.6980},
  year={2014}
}

@inproceedings{dlinear,
  title={Are transformers effective for time series forecasting?},
  author={Zeng, Ailing and Chen, Muxi and Zhang, Lei and Xu, Qiang},
  booktitle={Proceedings of the AAAI Conference on Artificial Intelligence},
  volume={37},
  pages={11121--11128},
  year={2023}
}

@inproceedings{
kim2022reversible,
title={Reversible Instance Normalization for Accurate Time-Series Forecasting against Distribution Shift},
author={Taesung Kim and Jinhee Kim and Yunwon Tae and Cheonbok Park and Jang-Ho Choi and Jaegul Choo},
booktitle={International Conference on Learning Representations},
year={2022}
}

@article{vaswani2017attention,
  title={Attention is all you need},
  author={Vaswani, Ashish and Shazeer, Noam and Parmar, Niki and Uszkoreit, Jakob and Jones, Llion and Gomez, Aidan N and Kaiser, {\L}ukasz and Polosukhin, Illia},
  journal={Advances in Neural Information Processing Systems},
  volume={30},
  year={2017}
}

@inproceedings{timesnet,
  title={Timesnet: Temporal 2d-variation modeling for general time series analysis},
  author={Wu, Haixu and Hu, Tengge and Liu, Yong and Zhou, Hang and Wang, Jianmin and Long, Mingsheng},
  booktitle={International Conference on Learning Representations},
  year={2022}
}

@article{autoformer,
  title={Autoformer: Decomposition transformers with auto-correlation for long-term series forecasting},
  author={Wu, Haixu and Xu, Jiehui and Wang, Jianmin and Long, Mingsheng},
  journal={Advances in Neural Information Processing Systems},
  volume={34},
  pages={22419--22430},
  year={2021}
}

@inproceedings{gu2021efficiently,
  title={Efficiently Modeling Long Sequences with Structured State Spaces},
  author={Gu, Albert and Goel, Karan and Re, Christopher},
  booktitle={International Conference on Learning Representations},
  year={2021}
}

@article{gu2023mamba,
  title={Mamba: Linear-time sequence modeling with selective state spaces},
  author={Gu, Albert and Dao, Tri},
  journal={arXiv preprint arXiv:2312.00752},
  year={2023}
}

@article{timemachine,
  title={{TimeMachine}: A time series is worth 4 mambas for long-term forecasting},
  author={Ahamed, Md Atik and Cheng, Qiang},
  journal={arXiv preprint arXiv:2403.09898},
  year={2024}
}

@article{menati2023modeling,
  title={Modeling and analysis of utilizing cryptocurrency mining for demand flexibility in electric energy systems: A synthetic texas grid case study},
  author={Menati, Ali and Lee, Kiyeob and Xie, Le},
  journal={IEEE Transactions on Energy Markets, Policy and Regulation},
  volume={1},
  number={1},
  pages={1--10},
  year={2023},
  publisher={IEEE}
}

@article{shi2024demand,
  title={Demand-side price-responsive flexibility and baseline estimation through end-to-end learning},
  author={Shi, Yuanyuan and Xu, Bolun},
  journal={IET Renewable Power Generation},
  volume={18},
  number={3},
  pages={361--371},
  year={2024},
  publisher={Wiley Online Library}
}

@article{menati2023high,
  title={High resolution modeling and analysis of cryptocurrency mining’s impact on power grids: Carbon footprint, reliability, and electricity price},
  author={Menati, Ali and Zheng, Xiangtian and Lee, Kiyeob and Shi, Ranyu and Du, Pengwei and Singh, Chanan and Xie, Le},
  journal={Advances in Applied Energy},
  volume={10},
  pages={100136},
  year={2023},
  publisher={Elsevier}
}

@ARTICLE{9732470,
  author={Kim, Nakyoung and Park, Hyunseo and Lee, Joohyung and Choi, Jun Kyun},
  journal={IEEE Transactions on Smart Grid}, 
  title={Short-Term Electrical Load Forecasting With Multidimensional Feature Extraction}, 
  year={2022},
  volume={13},
  number={4},
  pages={2999-3013},
  doi={10.1109/TSG.2022.3158387}}

@ARTICLE{10012043,
  author={Zhang, Chenxu and Fu, Yong},
  journal={IEEE Transactions on Power Systems}, 
  title={Probabilistic Electricity Price Forecast With Optimal Prediction Interval}, 
  year={2024},
  volume={39},
  number={1},
  pages={442-452},
  doi={10.1109/TPWRS.2023.3235193}}

@article{yang2017electricity,
  title={Electricity price forecasting by a hybrid model, combining wavelet transform, ARMA and kernel-based extreme learning machine methods},
  author={Yang, Zhang and Ce, Li and Lian, Li},
  journal={Applied Energy},
  volume={190},
  pages={291--305},
  year={2017},
  publisher={Elsevier}
}

@inproceedings{fu2023masked,
  title={Masked multi-step probabilistic forecasting for short-to-mid-term electricity demand},
  author={Fu, Yiwei and Virani, Nurali and Wang, Honggang},
  booktitle={2023 IEEE Power \& Energy Society General Meeting (PESGM)},
  pages={1--5},
  year={2023},
  organization={IEEE}
}

@ARTICLE{10111057,
  author={Rubasinghe, Osaka and Zhang, Xinan and Chau, Tat Kei and Chow, Yau Hing and Fernando, Tyrone and Iu, Herbert Ho-Ching},
  journal={IEEE Transactions on Power Systems}, 
  title={A Novel Sequence to Sequence Data Modelling Based CNN-LSTM Algorithm for Three Years Ahead Monthly Peak Load Forecasting}, 
  year={2024},
  volume={39},
  number={1},
  pages={1932-1947},
  doi={10.1109/TPWRS.2023.3271325}}

@article{tay2020long,
  title={Long range arena: A benchmark for efficient transformers},
  author={Tay, Yi and Dehghani, Mostafa and Abnar, Samira and Shen, Yikang and Bahri, Dara and Pham, Philip and Rao, Jinfeng and Yang, Liu and Ruder, Sebastian and Metzler, Donald},
  journal={arXiv preprint arXiv:2011.04006},
  year={2020}
}

@inproceedings{menati2022competitive,
  title={Competitive prediction-aware online algorithms for energy generation scheduling in microgrids},
  author={Menati, Ali and Chau, Sid Chi-Kin and Chen, Minghua},
  booktitle={Proceedings of the Thirteenth ACM International Conference on Future Energy Systems},
  pages={383--394},
  year={2022}
}

@inproceedings{haddad2019wind,
  title={Wind and solar forecasting for renewable energy system using sarima-based model},
  author={Haddad, Maroua and Nicod, Jean and Mainassara, Yacouba Boubacar and Rabehasaina, Landy and Al Masry, Zeina and P{\'e}ra, Marie},
  booktitle={International conference on time series and forecasting},
  year={2019}
}

@article{wang2020day,
  title={A day-ahead PV power forecasting method based on LSTM-RNN model and time correlation modification under partial daily pattern prediction framework},
  author={Wang, Fei and Xuan, Zhiming and Zhen, Zhao and Li, Kangping and Wang, Tieqiang and Shi, Min},
  journal={Energy Conversion and Management},
  volume={212},
  pages={112766},
  year={2020},
  publisher={Elsevier}
}

@article{jahangir2020deep,
  title={Deep learning-based forecasting approach in smart grids with microclustering and bidirectional LSTM network},
  author={Jahangir, Hamidreza and Tayarani, Hanif and Gougheri, Saleh Sadeghi and Golkar, Masoud Aliakbar and Ahmadian, Ali and Elkamel, Ali},
  journal={IEEE Transactions on Industrial Electronics},
  volume={68},
  number={9},
  pages={8298--8309},
  year={2020},
  publisher={IEEE}
}

@article{lara2020temporal,
  title={Temporal convolutional networks applied to energy-related time series forecasting},
  author={Lara-Ben{\'\i}tez, Pedro and Carranza-Garc{\'\i}a, Manuel and Luna-Romera, Jos{\'e} M and Riquelme, Jos{\'e} C},
  journal={applied sciences},
  volume={10},
  number={7},
  pages={2322},
  year={2020},
  publisher={MDPI}
}

@article{zuo2023ensemble,
  title={An ensemble framework for short-term load forecasting based on timesnet and tcn},
  author={Zuo, Chuanhui and Wang, Jialong and Liu, Mingping and Deng, Suhui and Wang, Qingnian},
  journal={Energies},
  volume={16},
  number={14},
  pages={5330},
  year={2023},
  publisher={MDPI}
}

@article{wang2023improved,
  title={An improved Wavenet network for multi-step-ahead wind energy forecasting},
  author={Wang, Yun and Chen, Tuo and Zhou, Shengchao and Zhang, Fan and Zou, Ruming and Hu, Qinghua},
  journal={Energy Conversion and Management},
  volume={278},
  pages={116709},
  year={2023},
  publisher={Elsevier}
}

@article{wu2022interpretable,
  title={Interpretable wind speed prediction with multivariate time series and temporal fusion transformers},
  author={Wu, Binrong and Wang, Lin and Zeng, Yu-Rong},
  journal={Energy},
  volume={252},
  pages={123990},
  year={2022},
  publisher={Elsevier}
}

@article{azam2021multi,
  title={Multi-horizon electricity load and price forecasting using an interpretable multi-head self-attention and EEMD-based framework},
  author={Azam, Muhammad Furqan and Younis, Muhammad Shahzad},
  journal={IEEE Access},
  volume={9},
  pages={85918--85932},
  year={2021},
  publisher={IEEE}
}

@article{hippert2001neural,
  title={Neural networks for short-term load forecasting: A review and evaluation},
  author={Hippert, Henrique Steinherz and Pedreira, Carlos Eduardo and Souza, Reinaldo Castro},
  journal={IEEE Transactions on power systems},
  volume={16},
  number={1},
  pages={44--55},
  year={2001},
  publisher={IEEE}
}

@article{xu2023power,
  title={Power-load forecasting model based on informer and its application},
  author={Xu, Hongbin and Peng, Qiang and Wang, Yuhao and Zhan, Zengwen},
  journal={Energies},
  volume={16},
  number={7},
  pages={3086},
  year={2023},
  publisher={MDPI}
}

@article{pombo2022benchmarking,
  title={Benchmarking physics-informed machine learning-based short term PV-power forecasting tools},
  author={Pombo, Daniel V{\'a}zquez and Bacher, Peder and Ziras, Charalampos and Bindner, Henrik W and Spataru, Sergiu V and S{\o}rensen, Poul E},
  journal={Energy Reports},
  volume={8},
  pages={6512--6520},
  year={2022},
  publisher={Elsevier}
}

@article{zheng2022multi,
  title={A multi-scale time-series dataset with benchmark for machine learning in decarbonized energy grids},
  author={Zheng, Xiangtian and Xu, Nan and Trinh, Loc and Wu, Dongqi and Huang, Tong and Sivaranjani, S and Liu, Yan and Xie, Le},
  journal={Scientific Data},
  volume={9},
  number={1},
  pages={359},
  year={2022},
  publisher={Nature Publishing Group UK London}
}

@book{abur2004state,
  author = {Abur, Ali and Gómez-Expósito, Antonio},
  title = {Power System State Estimation: Theory and Implementation},
  publisher = {Marcel Dekker},
  year = {2004}
}

@book{kundur1994stability,
  author = {Kundur, Prabha},
  title = {Power System Stability and Control},
  publisher = {McGraw-Hill Education},
  year = {1994}
}

@article{contreras2003arima,
  author = {Contreras, Javier and Espinola, Ricardo and Nogales, Francisco J. and Conejo, Antonio J.},
  title = {ARIMA Models to Predict Next-Day Electricity Prices},
  journal = {IEEE Transactions on Power Systems},
  volume = {18},
  number = {3},
  pages = {1014--1020},
  year = {2003},
  doi = {10.1109/TPWRS.2002.804943}
}

@ARTICLE{5667072,
  author={Dufour, Christian and Mahseredjian, Jean and Bélanger, Jean},
  journal={IEEE Transactions on Power Delivery}, 
  title={A Combined State-Space Nodal Method for the Simulation of Power System Transients}, 
  year={2011},
  volume={26},
  number={2},
  pages={928-935}}

@ARTICLE{4539776,
  author={Dehghani, Maryam and Nikravesh, Seyyed Kamaleddin Yadavar},
  journal={IEEE Transactions on Power Systems}, 
  title={State-Space Model Parameter Identification in Large-Scale Power Systems}, 
  year={2008},
  volume={23},
  number={3},
  pages={1449-1457}}

@article{hussain2023enhancing,
  title={Enhancing the efficiency of electric vehicles charging stations based on novel fuzzy integer linear programming},
  author={Hussain, Shahid and Irshad, Reyazur Rashid and Pallonetto, Fabiano and Jan, Qasim and Shukla, Saurabh and Thakur, Subhasis and Breslin, John G and Kim, Yun-Su and Rathore, Muhammad Ahmad and El-Sayed, Hesham},
  journal={IEEE Transactions on Intelligent Transportation Systems},
  volume={24},
  number={9},
  pages={9150--9164},
  year={2023},
  publisher={IEEE}
}

@article{hussain2022optimization,
  title={Optimization of waiting time for electric vehicles using a fuzzy inference system},
  author={Hussain, Shahid and Kim, Yun-Su and Thakur, Subhasis and Breslin, John G},
  journal={IEEE Transactions on Intelligent Transportation Systems},
  volume={23},
  number={9},
  pages={15396--15407},
  year={2022},
  publisher={IEEE}
}

@article{zhang2025unlocking,
  title={Unlocking the flexibilities of data centers for smart grid services: Optimal dispatch and design of energy storage systems under progressive loading},
  author={Zhang, Yingbo and Tang, Hong and Li, Hangxin and Wang, Shengwei},
  journal={Energy},
  volume={316},
  pages={134511},
  year={2025},
  publisher={Elsevier}
}

\begin{IEEEbiography}
[{\includegraphics[width=1in,height=1.25in,clip,keepaspectratio]{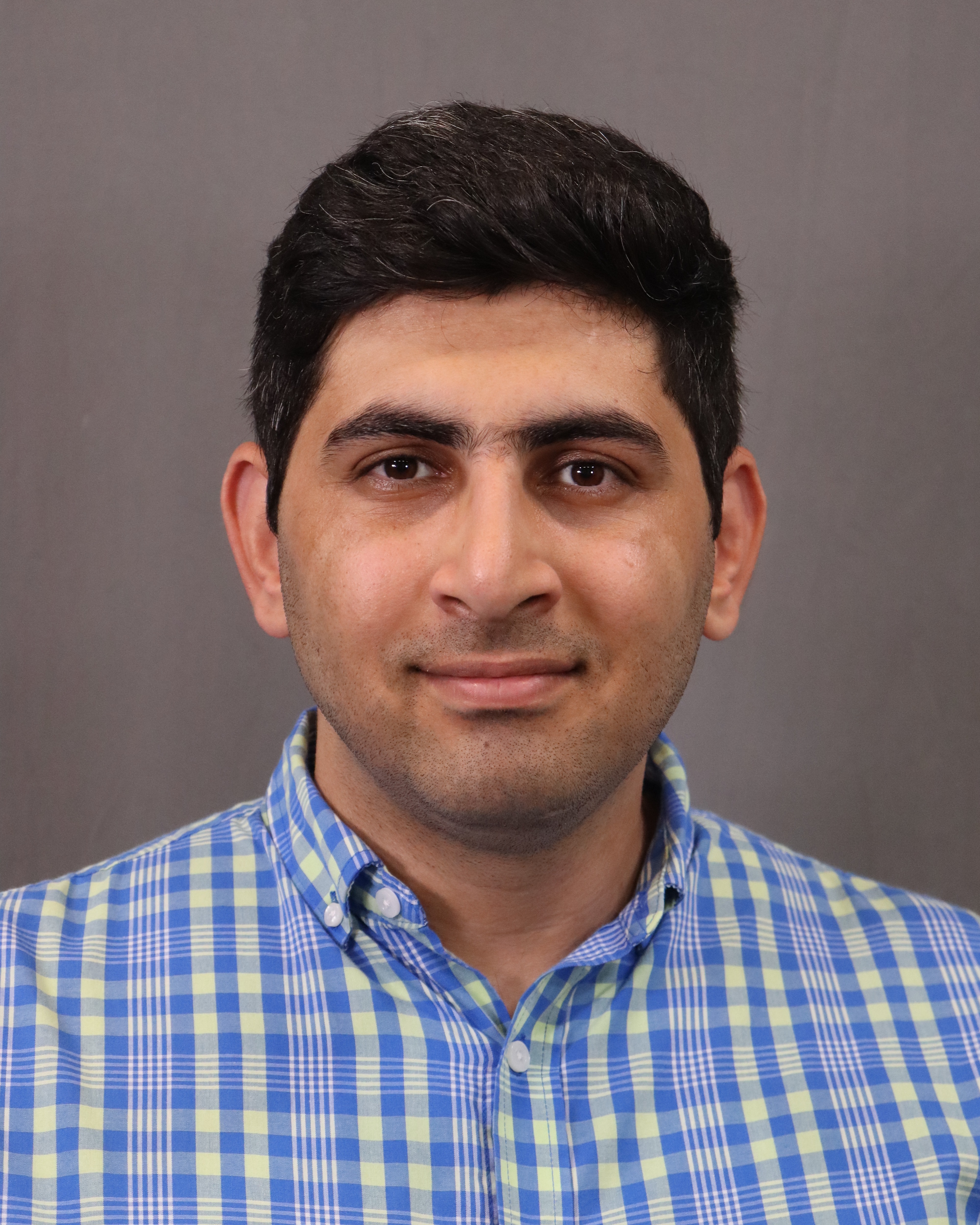}}]{Ali Menati}
received his B.Sc. degree in Electrical Engineering from Sharif University of Technology, Tehran, Iran, the M.Phil. degree in Information Engineering from the Chinese University of Hong Kong, and the Ph.D. degree in Electrical Engineering from Texas A\&M University, TX, USA. His research interests include online optimization and machine learning for electricity markets, as well as modeling and analysis of large flexible loads.
\end{IEEEbiography}

\begin{IEEEbiography}
[{\includegraphics[width=1in,height=1.25in,clip,keepaspectratio]{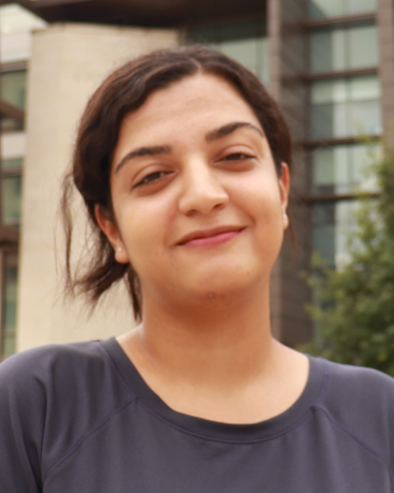}}]{Fatemeh Doudi}
received her B.Sc. and M.Sc. degrees from Sharif University of Technology, Tehran, Iran, and is currently a Ph.D. student in Computer Engineering at Texas A\&M University. Her research interests include generative models, with a focus on reinforcement learning--based alignment, inference-time scaling, and their application to domain-specific problems.
\end{IEEEbiography}

\begin{IEEEbiography}
[{\includegraphics[width=1in,height=1.25in,clip,keepaspectratio]{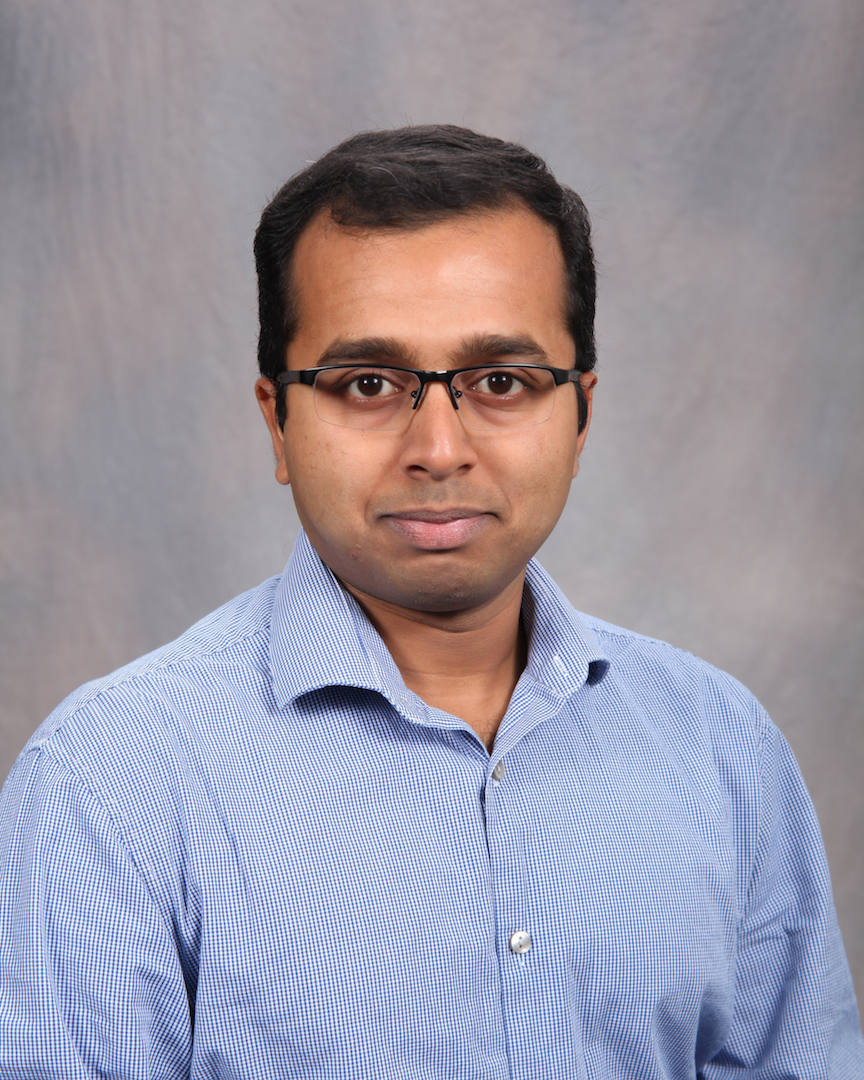}}]{Dileep Kalathil}(Senior Member, IEEE) received the Ph.D. degree from the University of Southern California in 2014. He was a Post-Doctoral Scholar with the University of California at Berkeley till 2017. He joined Texas A\&M University in 2017, where he is currently an Associate Professor of computer engineering with the Department of Electrical and Computer Engineering. Together with Srinivas Shakkottai, he co-directs the Learning and Emerging Networked Systems (LENS) Laboratory. His research interests include reinforcement learning, control theory, game theory and their applications in intelligent transportation systems, renewable energy systems, and cyber-physical systems. He has received the NSF CRII Award as well as the NSF Career Award in 2021.
\end{IEEEbiography}

\begin{IEEEbiography}
[{\includegraphics[width=1in,height=1.25in,clip,keepaspectratio]{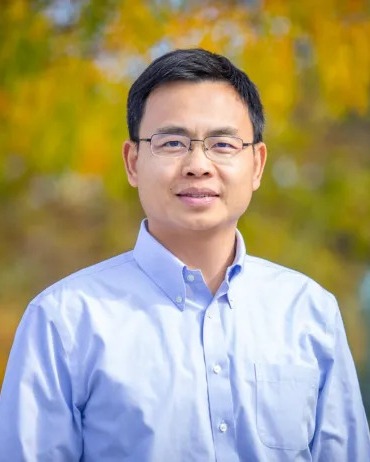}}]{Le Xie} (Fellow, IEEE) received the B.E. degree in electrical engineering from Tsinghua University, Beijing, the M.S. degree in engineering sciences from Harvard University, Cambridge, MA,  and the Ph.D. degree from Carnegie Mellon University, Pittsburgh, PA, USA. He is currently Gordon McKay Professor of Electrical Engineering at Harvard John A. Paulson School of Engineering and Applied Sciences. His research interests include modeling and control of large-scale complex systems, smart grids application with renewable energy resources, and electricity markets.
\end{IEEEbiography}

\end{document}